\documentclass{article}
\ifdefined\pdfminorversion \pdfminorversion=7 \fi
\PassOptionsToPackage{numbers, compress}{natbib}

\usepackage[eandd, final]{neurips_2026}

\PassOptionsToPackage{table}{xcolor}
\usepackage[utf8]{inputenc} \usepackage[T1]{fontenc} \usepackage{hyperref} \usepackage{url} \usepackage{booktabs} \usepackage{amsfonts} \usepackage{amsmath,amssymb} \usepackage{mathtools} \usepackage{nicefrac} \usepackage{microtype} \usepackage{xcolor} \usepackage{colortbl} \usepackage{graphicx} \usepackage{tabularx} \usepackage{multirow} \usepackage{subcaption} \usepackage{makecell} \usepackage{algorithm} \usepackage{algorithmic} \usepackage{xspace} \usepackage{float} \usepackage{placeins}
\usepackage{caption} \usepackage{enumitem}

    \newcommand{\etal}{\textit{et~al.}\xspace}

\newcommand{\Recog}{\textsc{Re:Cognize}\xspace} \newcommand{\PopChars}{\textsc{POPCharacters}\xspace} \newcommand{\Vouch}{\textsc{Vouch}\xspace} \newcommand{\MeCha}{\textsc{MeCha}\xspace}

\title{\Recog \ - A Framework For Open-Set \\Comic Character Re-Identification}

\author{%
  Aaditya Baranwal\textsuperscript{\dag}\quad Madhav Kataria\textsuperscript{\dag}\thanks{Work done as an intern at the University of Central Florida.\quad\textsuperscript{\dag}Equal contribution.}\hphantom{\textsuperscript{$\ast$}}\quad Yogesh S.\ Rawat\quad Shruti Vyas\\
  Institute of Artificial Intelligence, University of Central Florida, Orlando, FL, USA\\
  \texttt{aaditya.baranwal@ucf.edu}
}

\renewcommand{\topfraction}{0.9}
\renewcommand{\bottomfraction}{0.8}
\renewcommand{\textfraction}{0.07}
\renewcommand{\floatpagefraction}{0.75}
\newcommand{\blfootnote}[1]{\begingroup\renewcommand\thefootnote{}\footnote{#1}\addtocounter{footnote}{-1}\endgroup}

\begin{document}
\maketitle

\begin{center}\captionsetup{hypcap=false}\includegraphics[width=0.9\linewidth]{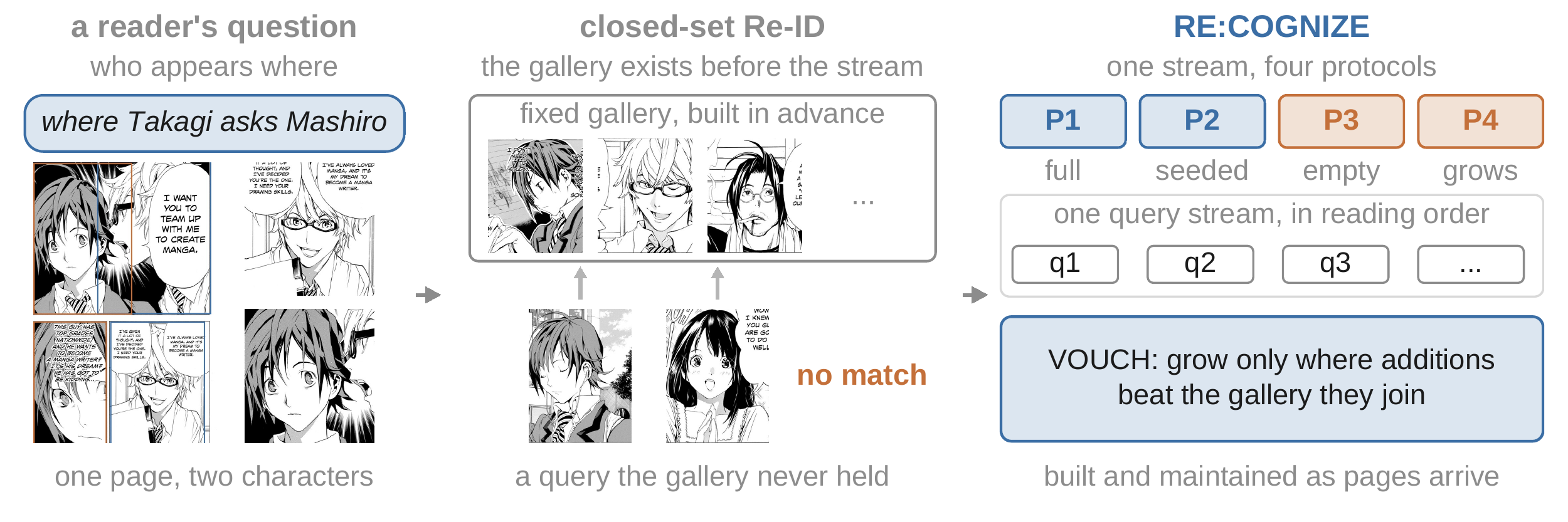}\captionof{figure}{\textbf{Recognising characters while the story is read.} On the left, a reader's question such as where Takagi asks Mashiro depends on who appears on which page. In the centre, standard Re-ID matches each query against a gallery built before reading starts, so a new character finds no match. On the right, \Recog streams crops in reading order against four galleries that hold every character (P1), a few labelled examples of each (P2), none (P3) or a few that grow as it reads (P4), and \Vouch grows the gallery only where the page vouches for a crop.}\label{fig:teaser}\end{center}\vspace{-8pt}

\begin{abstract}

Comics and manga tell stories through characters whose appearances change from panel to panel, yet readers recognise them hundreds of pages later without ever consulting a cast list. Character re-identification models rarely face the same challenge: existing evaluations give them a fixed set of identities before the story begins. \textbf{\Recog}\blfootnote{Code is available at \url{https://github.com/eternal-f1ame/Re-Cognize}.} instead asks models to recognise characters as the story unfolds and learn who is who along the way.
Across three manga corpora, we evaluate five Re-ID backbones and \textbf{\MeCha}, a memory-augmented baseline we introduce, and find that the central difficulty lies somewhere unexpected. Recognising familiar characters is within reach. Knowing which matches to trust is not. When models build a gallery and grow it from their own predictions, their performance deteriorates, even though correct identity labels would improve it by over twenty points. This raises a simple question: when should a model trust its own match?
We derive a single comparison that determines when gallery updates improve recognition and use it to design \Vouch, which updates its memory only when the page provides supporting evidence. Without fitting additional parameters, \Vouch recovers between one-third and two-thirds of the improvement available from perfect labels. \Recog measures whether a model can recognise characters as it reads; \Vouch shows how to remember them reliably. We focus our methodological claims on identity maintenance for previously encountered characters, while separately evaluating new-character emergence as a diagnostic under a fixed reference rule to distinguish recognition from discovery.

\end{abstract}

\section{Introduction} \label{sec:intro}

\begin{figure}[t]
\centering \includegraphics[width=\linewidth]{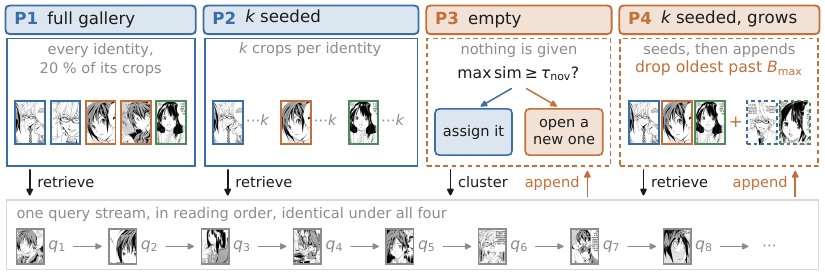} \caption{\textbf{The four \Recog protocols.} All four answer one stream of query crops in reading order and differ in the gallery and whether it may change. $k$ is the number of seed crops per character, $\tau_{\mathrm{nov}}$ the novelty threshold of P3 and $B_{\max}$ the cap on crops P4 adds per character. Frames are coloured by character and dashed where the model added the crop.} \label{fig:protocols}
\end{figure}

Manga and comics tell their stories over hundreds of pages, and a reader carries the cast from one page to the next. A face met in the first chapter is recognised in the tenth even when it is drawn from a new angle and in new clothes and is never named. Searching comics, indexing them, reasoning about their stories and making them accessible all rest on this skill, yet no system can reliably follow a character from one page to the next. We ask three simple questions.

\textbf{Can a model recognise a character it has met only once?} Re-identification (Re-ID) answers a version of this question in which every character is known in advance~\citep{ye2021deep,zheng2015scalable,zhang2022unsupervisedmangacharacterreidentification} and each query is matched against a gallery of reference crops built before the first query arrives. A volume read for the first time comes with no such gallery. New characters appear before anyone names them and their designs drift from arc to arc, so a model has to learn the cast while it reads.

\textbf{Can it build the cast as it reads?} A reader's picture of the cast grows with the story, and a model's gallery can grow too by filing each new crop under the character it seems to show. But a gallery is only as good as what it lets in, and a single crop filed under the wrong name keeps quietly attracting wrong matches page after page for as long as the story runs.

\textbf{When should a model believe its own match?} Every addition is a bet that the match is right. Confidence alone cannot tell whether the bet will pay, because the answer also depends on what the gallery already gets right, and knowing when to hold back matters as much as what to add.

To answer the three questions we introduce
\begin{itemize}[leftmargin=*,itemsep=1pt,topsep=2pt]
\item \textbf{\Recog}, the first framework that evaluates recognition as the story is read. One stream of crops in reading order meets four galleries (Figure~\ref{fig:protocols}). The closed set is P1 and a few labelled seeds per character are P2, while P3 starts empty for the model to organise into characters and in P4 the model grows its seeds as it reads. Our claims are on identity maintenance in P1, P2 and P4, while P3 measures emergence as a diagnostic under a fixed reference rule rather than a clustering algorithm we propose. We run it on three Japanese-comic corpora \PopChars~\citep{sachdeva2024tails}, Manga109~\citep{Aizawa_2020} and Re:Verse~\citep{baranwal2025reverse} with nearly 44{,}000 crops from 51 series and volumes, and test five Re-ID backbones and \MeCha, a memory-augmented maintenance baseline we introduce that gives any frozen backbone a working and an episodic memory of each character. A single reference image per character already matches closed-set retrieval, leaving gallery growth as the larger challenge. Growing a gallery from the model's own matches makes it worse on every backbone while correct labels would add over twenty points, and adapting the encoder does not close this gap. The limit is what the gallery agrees to accept as it reads (Section~\ref{sec:findings}). 
\item \textbf{The commit condition} characterises exactly when a gallery update improves recognition (Section~\ref{sec:condition}). An update pays when it identifies the queries it takes over more accurately than the original gallery. The comparison requires no fitted parameters, applies to three kinds of change (Figure~\ref{fig:one_breakeven}), and predicts their effects on held-out splits of each corpus.
\item \textbf{\Vouch}, which grows the gallery only where the page vouches for a crop (Section~\ref{sec:vouch}). It keeps one running average per character and adds a crop only when a crop grouped with it on its page is already filed under that character, and for first-appearance seeds a binding links each character's crops within and across pages (Section~\ref{sec:binding}). \Vouch improves on the unchanged gallery on \PopChars under both seeding regimes, with the largest gains from first-appearance seeds. On Manga109, the strongest gallery demonstrates why updating can sometimes hurt.
\end{itemize}

\section{Related work} \label{sec:related_work}


\begin{table}[t]
\centering \footnotesize \setlength{\tabcolsep}{4pt} \renewcommand{\arraystretch}{1.02} \caption{\textbf{Method positioning.} Prior comic and person Re-ID against \Recog\ across the six capabilities its protocols exercise. \emph{Memory} is the \MeCha baseline \Recog\ evaluates. \emph{Cross corpus} means evaluation on a second annotated corpus of the same medium.} \label{tab:positioning}
\begin{tabular}{l|c|c|c|c|c|c}
\toprule
\rowcolor{gray!15}
\textbf{Method} &
\makecell{\textbf{Closed} \textbf{Set}} &
\makecell{\textbf{Open}\\\textbf{Set}} &
\textbf{Sequential} &
\makecell{\textbf{Online}\\\textbf{Gallery}} &
\makecell{\textbf{Cross}\\\textbf{Corpus}} &
\textbf{Memory} \\
\midrule
TransReID~\citep{he2021transreid}         & \checkmark & -- & -- & -- & \checkmark & -- \\
OSNet~\citep{zhou2019osnet}               & \checkmark & -- & -- & -- & \checkmark & -- \\
Instruct-ReID~\citep{he2023instructreid}  & \checkmark & -- & -- & -- & \checkmark & -- \\
Zhang \etal~\citep{zhang2022unsupervisedmangacharacterreidentification} & \checkmark & -- & \checkmark & -- & -- & -- \\
Zhang \& Chu~\citep{zhang2023occlusion}   & \checkmark & -- & -- & -- & -- & -- \\
Soykan \etal~\citep{soykan2023identity}   & \checkmark & -- & -- & -- & -- & -- \\
Video Re-ID~\citep{wu2016deep}            & \checkmark & -- & \checkmark & -- & -- & -- \\
Lifelong Re-ID~\citep{liu2025dafc}        & \checkmark & -- & -- & -- & \checkmark & -- \\
\midrule
\rowcolor{gray!15}
\textbf{\Recog (Ours)} & \checkmark & \checkmark & \checkmark & \checkmark & \checkmark & \checkmark \\
\bottomrule
\end{tabular}
\end{table}

\textbf{Comic character analysis, and person and open-set Re-ID.} Comic understanding has moved from panel detection~\citep{ogawa2018object} to character analysis on Manga109~\citep{Aizawa_2020}, to manga-native Re-ID models~\citep{sachdeva2024manga,sachdeva2024tails} and to character Re-ID through clustering~\citep{zhang2022unsupervisedmangacharacterreidentification,zhang2023occlusion} and semi-supervised association~\citep{soykan2023identity}. Person Re-ID is closed-set by construction~\citep{ye2021deep,zheng2015scalable,he2021transreid,luo2019bag,hermans2017defensetripletlossperson,dosovitskiy2020vit}, and each line beyond it relaxes one assumption while keeping the rest. Video Re-ID~\citep{wu2016deep,mclaughlin2016recurrent} adds context within a tracklet, domain-adaptive Re-ID~\citep{zheng2021exploiting,ge2020selfpaced} crosses corpora, open-set recognition~\citep{scheirer2013toward,wang2016opensetreid} scores rejection, lifelong Re-ID~\citep{liu2025dafc,pu2021lifelong} accumulates over known task boundaries and instruction-conditioned retrieval~\citep{he2023instructreid} changes the query. All of them fix the cast or the gallery in advance, and none scores sequential processing, online gallery construction and unknown identities at once.

\textbf{Gallery representation.} Representing a class by the mean of its examples is standard~\citep{snell2017prototypical,vinyals2016matching,deng2019arcface}, and \Vouch's per-character average is exactly that. These methods fix their support set in advance and so never need a rule for when averaging helps a gallery that is still being built. On a stream the answer depends on how accurate the gallery already is, which Section~\ref{sec:condition} makes precise.

\textbf{Self-training, pseudo-labels and template drift.} A gallery grown by its own matches learns from labels it assigned itself. Pseudo-labelling does the same and suffers confirmation bias~\citep{lee2013pseudo,arazo2020pseudo}, a tracker updated with its own matches drifts~\citep{matthews2004template}, test-time adaptation accumulates its own errors~\citep{wang2021tent,wang2022cotta} and unsupervised Re-ID refines the cluster labels it trains on~\citep{ge2020mutual,ge2020selfpaced}. The usual remedy gates each sample by confidence through a fixed threshold~\citep{sohn2020fixmatch}, an uncertainty estimate~\citep{rizve2021ups} or a quantity-quality weighting~\citep{chen2023softmatch}, and analyses of self-training reason from the labeller's error~\citep{wei2021theoretical,kalal2012tld}. The commit condition instead weighs the labeller against the gallery it would change, so the same labeller can help a weak gallery and harm a strong one, which no per-sample threshold can express. Confidence alone would not separate them here anyway, since right and wrong top-1 matches alike score above a cosine similarity of 0.7 on every backbone (Appendix~\ref{app:acceptance}).

\textbf{Long-form narrative and memory.} Vision-language models~\citep{radford2021clip,liu2024llava,bai2023qwen} lose track of who is who across long sequences~\citep{wang2025beyondsingleframes,vivoli2025missingpiecevisionlanguage}, and Re:Verse~\citep{baranwal2025reverse} reports near-zero character identification on chapter-length manga. Memory and prototype models~\citep{snell2017prototypical,santoro2016meta,graves2014neural,sukhbaatar2015endtoend} carry context forward, and \MeCha instantiates that line. Table~\ref{tab:positioning} lists the six axes \Recog is the first to score together.

\textbf{What the protocols inherit from these lines.} P1 is the closed-set retrieval of person Re-ID benchmarks~\citep{zheng2015scalable,ye2021deep}, and P2 carries the few-shot support sets of prototypical and matching networks~\citep{snell2017prototypical,vinyals2016matching} onto a stream. P3 makes the novelty decision that open-set recognition evaluates through rejection~\citep{scheirer2013toward} and open-set Re-ID through membership in a fixed gallery~\citep{wang2016opensetreid}, but starts empty and serves as a diagnostic (Section~\ref{subsec:emergence}). P4 is the gallery growth of lifelong Re-ID~\citep{pu2021lifelong,liu2025dafc} without announced task boundaries. No earlier line scores one stream under all four as Table~\ref{tab:cross_protocol} does.
\section{The \Recog framework} \label{sec:task}

\begin{table}[t]
\centering
\footnotesize
\setlength{\tabcolsep}{4pt}
\renewcommand{\arraystretch}{1.0}
\caption{\textbf{Cross-protocol summary on \PopChars}, 8 held-out series, $k=1$ seed per character, mean over three training runs. \emph{Chance} is a random ranking of the same galleries. Per backbone the top row, \emph{Finetuned}, is the frozen backbone with a trained BNNeck and no LoRA, and the bottom row ($\dagger$) the best \MeCha configuration by P1 mAP. Seeds are random (Seq-R) or first appearances (Seq-T). P4 is identity Rank-1 at $B_{\max}=50$ under three policies, where \emph{static} keeps the P2 gallery while \emph{predicted} and \emph{oracle} add each query under its top-1 match and its true character.}
\label{tab:cross_protocol}
\resizebox{\textwidth}{!}{%
\begin{tabular}{l|l|cc|cc|ccc|ccc}
\toprule
\rowcolor{gray!15}
& & \multicolumn{2}{c|}{\textbf{P1: Closed-Set}} & \multicolumn{2}{c|}{\textbf{P2: Seeded}$_{k=1}$, mAP} & \multicolumn{3}{c|}{\textbf{P4: Seq-R}$_{k=1}$, id.\ R-1} & \multicolumn{3}{c}{\textbf{P4: Seq-T}$_{k=1}$, id.\ R-1} \\
\rowcolor{gray!15}
\multirow{-2}{*}{\textbf{Backbone}} & \multirow{-2}{*}{\textbf{Config}} & \textbf{mAP} & \textbf{R-1} & \textbf{Seq-R} & \textbf{Seq-T} & \textbf{Static} & \textbf{Pred.} & \textbf{Oracle} & \textbf{Static} & \textbf{Pred.} & \textbf{Oracle} \\
\midrule
\emph{Chance} & Random ranking & 33.1 & 30.1 & 33.9 & 33.9 & 12.9 & -- & -- & 12.9 & -- & -- \\
\midrule
TransReID & Finetuned & 37.4 & 40.6 & 38.6 & 33.2 & 17.4 & 15.0 & 41.7 & 12.8 & 12.6 & 42.2 \\
 & FT + Mem + LoRA$^\dagger$ & 38.1 & 39.9 & 39.0 & 34.2 & 18.1 & 16.1 & 42.3 & 13.9 & 13.3 & 42.4 \\
\midrule
MagiV2 & Finetuned & 51.2 & 57.2 & 54.4 & 52.9 & 36.8 & 36.3 & 59.1 & 36.6 & 37.9 & 59.0 \\
 & FT + Mem$^\dagger$ & 51.7 & 56.3 & 54.6 & 51.3 & 37.2 & 35.9 & 59.8 & 35.3 & 36.9 & 60.4 \\
\midrule
MagiV3 & Finetuned & 41.3 & 48.7 & 43.6 & 37.5 & 22.7 & 19.9 & 49.9 & 17.4 & 17.7 & 50.2 \\
 & FT + Mem$^\dagger$ & 42.4 & 47.6 & 44.4 & 38.9 & 23.5 & 19.6 & 50.5 & 17.7 & 17.5 & 50.7 \\
\midrule
InstructReID & Finetuned & 37.8 & 42.5 & 38.7 & 33.7 & 17.5 & 14.5 & 42.6 & 13.3 & 10.6 & 42.9 \\
 & FT + Mem + LoRA$^\dagger$ & 40.1 & 43.9 & 40.4 & 35.9 & 18.9 & 15.6 & 47.3 & 15.1 & 14.2 & 47.5 \\
\midrule
ReID5o & Finetuned & 38.7 & 44.8 & 41.6 & 33.9 & 20.8 & 17.9 & 45.6 & 12.8 & 14.4 & 45.6 \\
 & FT + Mem + LoRA$^\dagger$ & 41.0 & 46.9 & 43.7 & 36.3 & 23.1 & 19.1 & 48.5 & 14.4 & 16.7 & 48.5 \\
\bottomrule
\end{tabular}%
}
\end{table}

\textbf{Streaming task and two sub-problems.}\label{subsec:problem} A system reads a volume as a stream of character crops $\mathcal{S}=\{x_t\}_{t=1}^{T}$ in reading order. It keeps a \emph{gallery} of reference crops filed under known characters, and each new crop, the \emph{query}, must be named from it or recognised as new. Two sub-problems are coupled, \emph{identity emergence} for a character never seen and \emph{identity maintenance} for one already met. Maintenance is most of the stream, and the four protocols (Figure~\ref{fig:protocols}) measure the two apart.

\label{subsec:protocols}\textbf{P1 (closed-set retrieval).} A fifth of each character's crops form the gallery and the rest are queries, ranked by cosine similarity and scored by mAP and Rank-$k$. P1 is the closed-set ceiling.

\begin{figure}[t]
\centering
\begin{subfigure}[b]{0.48\linewidth}
\centering \includegraphics[width=\linewidth]{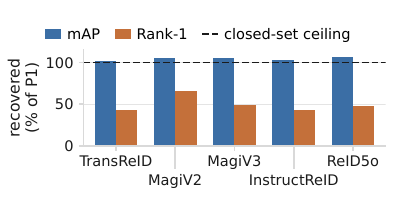} \caption{What one seed recovers, by metric.} \label{fig:ceiling_recovery}
\end{subfigure}\hfill
\begin{subfigure}[b]{0.48\linewidth}
\centering \includegraphics[width=\linewidth]{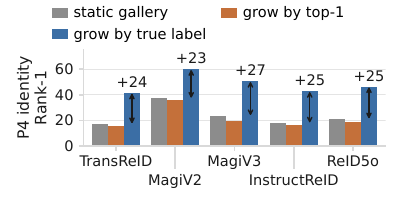} \caption{What correct growth would add, $k{=}1$.} \label{fig:p4_closes_gap}
\end{subfigure}\\[2pt]
\begin{subfigure}[b]{0.48\linewidth}
\centering \includegraphics[width=\linewidth]{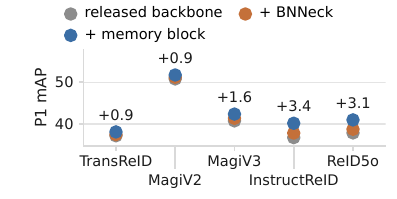} \caption{Encoder-side adaptation, three training runs.} \label{fig:adaptation}
\end{subfigure}\hfill
\begin{subfigure}[b]{0.48\linewidth}
\centering \includegraphics[width=\linewidth]{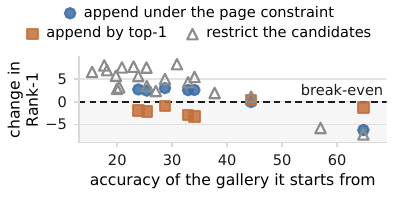} \caption{Every gallery operation at $k{=}5$, against the gallery it starts from.} \label{fig:one_breakeven}
\end{subfigure}
\caption{\textbf{What the protocols measure, and where the headroom is.} (a) One seed per character recovers the closed-set mAP but only part of its Rank-1. (b) Growth by the model's own top-1 matches stays below a static gallery, while correct labels would lift it far above (labels give oracle minus static). Panels (a) and (b) use \MeCha with one random seed per character over three training runs. (c) Encoder-side adaptation helps the backbones weakest in this domain most. (d) Each point is one change to the gallery on one backbone and corpus, whether adding under the page constraint (Section~\ref{sec:vouch}), adding by top-1 match or restricting the candidate characters. The stronger the gallery already is, the less any change adds, and Section~\ref{sec:condition} predicts which points fall below zero.} \label{fig:ceiling_and_gap}
\end{figure}

\textbf{P2 (seeded static gallery).} The gallery holds $k$ labelled crops per character, the \emph{seeds}, for $k$ from 1 to 5, and every other crop is a query. The seeds are drawn at random (\emph{random seeding}, Seq-R) or are the character's first $k$ appearances (\emph{chronological seeding}, Seq-T), the way a reader meets them.

\textbf{P3 (unsupervised online clustering).} Crops arrive unlabelled into an empty gallery and either join the nearest cluster above a novelty threshold $\tau_{\mathrm{nov}}$ or open a new one, and the result is scored by Purity, NMI and ARI. P3 is a measurement instrument, not a method. One fixed threshold compares every representation by one criterion, no claim depends on it being an optimal clustering algorithm, and four alternative rules only trade purity for fragmentation on one frontier (Appendix~\ref{app:p3}).

\textbf{P4 (seeded gallery that grows).} P4 starts from P2's gallery and, in reading order, adds each query under the character of its top-1 match. The seeds are kept and the oldest additions beyond $B_{\max}=50$ per character are replaced, and results barely move above a cap of 25 (Appendix~\ref{app:p4_dynamics}). Three policies separate what growth gives. \emph{Static} keeps the P2 gallery, \emph{predicted} adds each query under its top-1 match and \emph{oracle} adds each query under its true character. P4 is scored by \emph{identity Rank-1}, the share of queries whose top-ranked character is correct, which stays comparable as the gallery grows.

\textbf{What the protocols are built to measure.}\label{subsec:axes} Two comparisons read the protocols. The \emph{seeding} comparison of P1 with P2 asks how much of a closed-set gallery a few seeds can replace, and the \emph{acceptance} comparison of P4's three policies at the same $k$ asks what the stream can add and what a wrong addition costs. A weak representation would depress both, so when the two move apart the deficit lies in the gallery rather than in the encoder. Two floors anchor every number. A random ranking of the same galleries scores about 33 mAP at P1 and P2, because a few protagonists account for most crops, and 12.9 identity Rank-1 with one seed per character (Table~\ref{tab:cross_protocol}). A second training run typically moves mAP by well under a point and identity Rank-1 by under one and a half (Appendix~\ref{sec:validity}), and every headline comparison is paired over three training runs.

\textbf{Corpora, reference backbones and the \MeCha baseline.}\label{subsec:dataset}\label{subsec:mecha} The protocols run on \PopChars~\citep{sachdeva2024tails} (23 series with 13 for training, 2 for development and 8 held out, 12{,}599 crops), Manga109~\citep{Aizawa_2020} (27 held-out volumes and 29{,}315 crops, scored zero-shot) and Re:Verse~\citep{baranwal2025reverse} (one series, Re:Zero), each heavy-tailed per character (Appendix~\ref{app:per_series}). Five backbones span three regimes, person Re-ID with TransReID~\citep{he2021transreid} and the instruction-conditioned InstructReID~\citep{he2023instructreid}, the manga-native MagiV2~\citep{sachdeva2024manga} and MagiV3~\citep{sachdeva2024tails}, and ReID5o~\citep{zuo2025reid5o} built on CLIP. Each runs at its native feature width and is scored as released, with a fine-tuned batch-normalisation neck (BNNeck)~\citep{luo2019bag}, with LoRA~\citep{hu2021lora} and with \MeCha, whose working and episodic memories are fused into the frozen backbone through one gated residual and trained with a two-pass episodic protocol (Appendix~\ref{app:mecha_arch}). \MeCha places memory in the representation, and Section~\ref{sec:vouch} places it in the gallery.
\section{What the protocols measure} \label{sec:findings}

\textbf{A small gallery nearly matches closed-set retrieval, but growing it remains difficult.}\label{subsec:ceiling_and_growth} On the 8 held-out \PopChars series, with 70 characters and 4{,}058 crops, one seed per character is enough to rank (Table~\ref{tab:cross_protocol}). With \MeCha, P2 at $k=1$ reaches 102 to 107 percent of the P1 mAP ceiling on all five backbones and 43 to 66 percent of its Rank-1 (Figure~\ref{fig:ceiling_recovery}). It can pass the ceiling because the two galleries hold different numbers of entries per character and mAP rewards the smaller (Appendix~\ref{sec:validity}). Growth is where the gallery falls short. From one random seed per character, adding each query under its top-1 match lowers identity Rank-1 in all ten rows of Table~\ref{tab:cross_protocol}, while an oracle that adds every query under its true character raises it by 22 to 27 points, the largest effect we measure (Figure~\ref{fig:p4_closes_gap}). The cause is contamination, since 62 to 89 percent of the model's additions carry the wrong character (Appendix Table~\ref{tab:p4_oracle}). The loss holds steady through every quarter of the stream (Appendix~\ref{app:p4_dynamics}), a problem of acceptance rather than accumulation.

\begin{table}[t]
\centering \footnotesize \setlength{\tabcolsep}{5pt} \renewcommand{\arraystretch}{1.0} \caption{\textbf{P3 identity emergence on full per-series streams} (\PopChars test series, finetuned backbones from the first training run, macro over 8 series). The fixed rule at $\tau_{\mathrm{nov}}=0.55$ is the reference instantiation. The loose rows show why Purity cannot be read alone. Raising the threshold multiplies clusters four- to fivefold and Purity rises with them, while Hungarian accuracy and ARI collapse. Alternative decision rules are in Appendix Table~\ref{tab:suppl_p3_rules} and the comparison with \MeCha in Appendix Table~\ref{tab:suppl_p3}, and neither is a baseline of the framework.} \label{tab:p3_main}
\begin{tabular}{l|l|ccccc}
\toprule
\rowcolor{gray!15}
\textbf{Backbone} & \textbf{Rule} & \textbf{\#clusters} & \textbf{Purity} & \textbf{Hung.\ Acc} & \textbf{ARI} & \textbf{NMI} \\
\midrule
TransReID & fixed $\tau_{\mathrm{nov}}=0.55$ & 95.8 & 60.2 & 15.1 & 2.0 & 21.8 \\
MagiV2    & fixed $\tau_{\mathrm{nov}}=0.55$ & 38.9 & 69.6 & 46.6 & 24.0 & 34.1 \\
\midrule
TransReID & fixed $\tau_{\mathrm{nov}}=0.80$ (loose) & 431.1 & 93.0 & 4.8 & 0.2 & 37.0 \\
MagiV2    & fixed $\tau_{\mathrm{nov}}=0.80$ (loose) & 190.9 & 84.3 & 24.6 & 9.8 & 38.5 \\
\bottomrule
\end{tabular}
\end{table}

\begin{table}[t]
\centering
\footnotesize
\setlength{\tabcolsep}{4pt}
\renewcommand{\arraystretch}{1.0}
\caption{\textbf{The commit condition reproduces every measured change.} Commitment under the page constraint (Section~\ref{sec:vouch}) on the bag of $k=5$ random seeds per character, without the cast sheet, over three seed draws. $a$ is the static gallery's identity Rank-1 and $c$, $p_{\mathrm{eff}}$ and $a^{+}$ are the terms of Equation~\ref{eq:commit}, in percent. Weighting $p_{\mathrm{eff}}$ and $a^{+}$ by each record's capture rate, $\sum_i c_i x_i/\sum_i c_i$, makes $c\,(p_{\mathrm{eff}}-a^{+})$ exactly the mean measured change, so the last two columns agree.}
\label{tab:commit}
\begin{tabular}{ll|cccc|cc}
\toprule
\rowcolor{gray!15}
\textbf{Corpus} & \textbf{Backbone} & $\boldsymbol{a}$ & $\boldsymbol{c}$ & $\boldsymbol{p_{\mathrm{eff}}}$ & $\boldsymbol{a^{+}}$ & $\boldsymbol{c(p_{\mathrm{eff}}\!-\!a^{+})}$ & \textbf{measured} \\
\midrule
\PopChars & TransReID & 23.82 & 34.96\% & 31.532 & 23.791 & +2.71 & +2.71 \\
\PopChars & InstructReID & 25.38 & 33.15\% & 30.851 & 23.437 & +2.46 & +2.46 \\
\PopChars & ReID5o & 28.68 & 35.14\% & 36.682 & 28.106 & +3.01 & +3.01 \\
\PopChars & MagiV3 & 32.88 & 35.67\% & 39.119 & 31.918 & +2.57 & +2.57 \\
Manga109 & MagiV3 & 34.03 & 43.35\% & 39.513 & 33.519 & +2.60 & +2.60 \\
\PopChars & MagiV2 & 44.35 & 34.67\% & 47.832 & 47.960 & -0.04 & -0.04 \\
Manga109 & MagiV2 & 64.77 & 46.53\% & 54.418 & 67.799 & -6.23 & -6.23 \\
\bottomrule
\end{tabular}
\end{table}

\textbf{The encoder is not the bottleneck.}\label{subsec:profiles_ablations} Fine-tuning a BNNeck adds about a point of P1 mAP and the best configuration at most 3.4 (Figure~\ref{fig:adaptation}), with LoRA helping only the backbones not built for manga (Appendix Table~\ref{tab:suppl_p1_full}). \MeCha adds at most a point, all of it from working memory, whose removal significantly hurts both manga-native backbones (Appendix~\ref{app:ablation}).

\textbf{Transfer within Japanese comics.}\label{subsec:transfer}\label{subsec:cross_dataset_p3} Checkpoints trained on \PopChars\ transfer without retraining. On 27 held-out Manga109 volumes the best \MeCha configuration adds 1.2 to 5.1 P1 mAP on all five backbones. The manga-native backbones transfer best, and MagiV2 even scores higher there than in domain, while the other three lose several points (Appendix Tables~\ref{tab:manga109_results} and~\ref{tab:reverse_results}). On Re:Verse released MagiV2 reaches 84.2 P1 mAP, while three vision-language models asked to name the characters are right at most 1.1 percent of the time~\citep{baranwal2025reverse}.

\textbf{Emergence, measured as a diagnostic.}\label{subsec:emergence} P3 is a diagnostic and nothing is optimised against it. On full per-series streams at $\tau_{\mathrm{nov}}=0.55$, fine-tuned TransReID and MagiV2 open 95.8 and 38.9 clusters for 8.8 characters per series (Table~\ref{tab:p3_main}, Appendix~\ref{app:p3}). The same rule reaches an order of magnitude higher ARI on the comic-native representation, so emergence is bounded by the embedding's geometry rather than by the decision rule and is a question for representation learning. A looser threshold raises Purity only by splitting each character into dozens of clusters, which is why P3 reports ARI and the predicted cluster count beside it.
\section{When a change to the gallery pays} \label{sec:condition} \label{subsec:commit}

\begin{figure}[t]
\centering
\includegraphics[width=\linewidth]{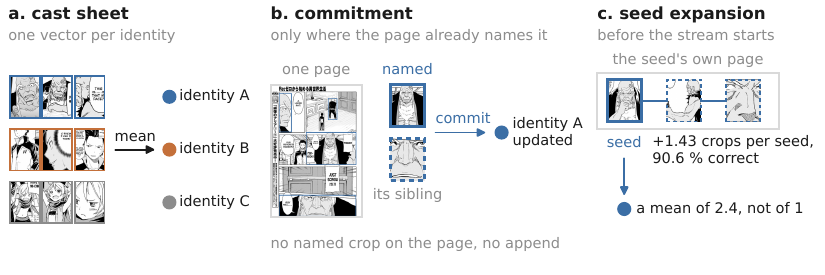}\\[2pt]
\includegraphics[width=\linewidth]{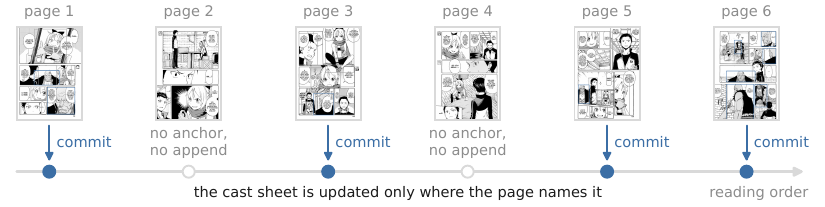}
\caption{\textbf{\Vouch} on Re:Zero crops from Re:Verse~\citep{baranwal2025reverse}, with Rom as identity A. The top row shows the three changes of Section~\ref{sec:vouch}. A page \emph{names} a character when one of its crops is already committed to it, and commitment then adds that crop's page-group sibling. In the bottom row, over six pages the cast sheet is updated only on the four that name Rom, and elsewhere the rule abstains.}
\label{fig:vouch_schematic}
\end{figure}

The gap of Section~\ref{subsec:ceiling_and_growth} is an acceptance gap. The correct crops are in the stream, and what fails is the rule that admits them. A change $A$ to the gallery can alter the answer only for a query whose nearest entry $A$ supplied. We call these queries \emph{captured} and write $Q_A$ for them within the stream's queries $Q$. With the capture rate $c=|Q_A|/|Q|$, $p_{\mathrm{eff}}$ the share of captured queries whose capturing entry carries their own identity and $a^{+}$ the accuracy the unchanged gallery would have had on them, the change in accuracy is
\begin{equation}
\Delta \;=\; c\,\bigl(p_{\mathrm{eff}} - a^{+}\bigr).
\label{eq:commit}
\end{equation}
The decomposition is exact (Table~\ref{tab:commit}), so a change pays exactly when $p_{\mathrm{eff}}>a^{+}$, which we call the \emph{commit condition}. Our running example is the page constraint of Section~\ref{sec:vouch}, which adds a crop to a character only when a crop grouped with it on its page is already filed under that character. Its additions are correct 86.9 percent of the time on \PopChars\ whatever the backbone.

\textbf{Both terms differ from the intuition.} Intuition compares how often the added crops are correct with the gallery's average accuracy, and neither is the right term. The precision that governs is $p_{\mathrm{eff}}$, only 31 to 48 percent on \PopChars, because a crop filed under the right character still takes over other characters' queries. The captured queries also lie close to crops of the stream where the gallery does well, so $a^{+}$ exceeds the average accuracy $a$ wherever the gallery is strong.

\textbf{The exception is the strongest gallery.} MagiV2's gallery on \PopChars\ is right 44.4 percent of the time, and the page constraint still gains nothing. Its additions are right about 47.8 percent of the queries they take over, and the gallery would have answered 48.0 percent of those correctly. On Manga109 MagiV2 starts at 64.8, a $p_{\mathrm{eff}}$ of 54.4 falls below an $a^{+}$ of 67.8, and growth loses 6.23 points exactly as Equation~\ref{eq:commit} gives. The intuitive test predicts a gain in both cases, whereas the three terms, measured on a labelled half of a corpus, make the right call for the other half in every random split wherever the change matters (Appendix~\ref{sec:validity}).

\textbf{The comparison is not specific to growth.} Restricting each query's candidates to the characters of the last twenty crops adds no crops at all, yet it gains 3.4 to 8.0 points of identity Rank-1 on four of five backbones across both corpora. It costs MagiV2 7.2 points on Manga109, where its gallery is strongest. A decomposition of the same shape explains why, since the cost grows with the gallery's accuracy and the benefit does not (Appendix~\ref{app:seqt}).
\section{\Vouch, three changes to the gallery} \label{sec:vouch}

\begin{table}[t]
\centering
\footnotesize
\setlength{\tabcolsep}{4pt}
\renewcommand{\arraystretch}{1.0}
\caption{\textbf{\Vouch, one change at a time.} P4 identity Rank-1 with random seeds at $B_{\max}=50$, on the 8 held-out \PopChars series and, zero-shot, on 27 held-out Manga109 volumes. Each column after \emph{Static} adds one change and gives the cumulative gain over \emph{Static}, except \emph{$+$ expansion}, which is scored against its own static reference with the expanded crops removed from the queries. \emph{Oracle} adds every query under its true label. Cells are one training run and three seed draws, and \textbf{bold} marks $p<0.05$ on a paired $t$-test over series. At $k=1$ the first two changes coincide, the oracle is in Table~\ref{tab:cross_protocol}, and seed expansion was not run on Manga109.}
\label{tab:vouch_combined}
\begin{tabular}{l|c|cc|c|c}
\toprule
\rowcolor{gray!15}
\textbf{Backbone} & \textbf{Static} & \textbf{Cast sheet} & \textbf{$+$ commitment} & \textbf{$+$ expansion} & \textbf{Oracle} \\
\midrule
\multicolumn{6}{l}{\PopChars\textit{, 8 test series, Seq-R, $k=1$}} \\
TransReID & 17.17 & \multicolumn{2}{c|}{+0.00} & +4.71 & --- \\
MagiV2 & 35.95 & \multicolumn{2}{c|}{+0.00} & +0.62 & --- \\
MagiV3 & 21.79 & \multicolumn{2}{c|}{+0.00} & +4.84 & --- \\
InstructReID & 16.83 & \multicolumn{2}{c|}{+0.00} & \textbf{+6.84} & --- \\
ReID5o & 21.03 & \multicolumn{2}{c|}{+0.00} & \textbf{+5.66} & --- \\
\midrule
\multicolumn{6}{l}{\PopChars\textit{, 8 test series, Seq-R, $k=5$}} \\
TransReID & 23.82 & +2.86 & \textbf{+4.27} & \textbf{+5.39} & +10.74 \\
MagiV2 & 44.35 & \textbf{+6.81} & \textbf{+6.51} & +5.58 & +15.97 \\
MagiV3 & 32.88 & +1.47 & \textbf{+3.34} & \textbf{+4.77} & +11.83 \\
InstructReID & 25.38 & +1.49 & \textbf{+3.74} & \textbf{+5.28} & +9.66 \\
ReID5o & 28.68 & +1.71 & \textbf{+3.74} & \textbf{+3.89} & +10.47 \\
\midrule
\multicolumn{6}{l}{\textit{Manga109, 27 held-out volumes, zero-shot, Seq-R, $k=5$}} \\
MagiV2 & 64.77 & \textbf{+2.78} & +1.33 & --- & +11.09 \\
MagiV3 & 34.03 & \textbf{+3.81} & \textbf{+6.20} & --- & +15.86 \\
\bottomrule
\end{tabular}
\end{table}

\label{subsec:axis}The commit condition turns growth into a design problem. Raise $p_{\mathrm{eff}}$, and act only where the additions will be right more often than the gallery they join. \Vouch\ (Figure~\ref{fig:vouch_schematic}) makes three changes to the gallery and fits nothing on data. Two act on the terms of Equation~\ref{eq:commit} and the third on what the seeds carry into the stream, and Section~\ref{sec:binding} adds a binding for first-appearance seeds.

\textbf{A cast sheet instead of a bag of crops.} Each character is one $\ell_2$-normalised average of the crops filed under it rather than a set of separate exemplars. A new crop then refines its character's entry instead of becoming a rival entry that takes over other characters' queries, which acts on the capture rate $c$. \textbf{Commitment under the page constraint.} A crop joins a character only when another crop in its page group is already committed to that character. The page groups come from the character-to-character affinity head MagiV2 uses for transcription~\citep{sachdeva2024tails}, run once per page with no training, and two crops it groups show the same character $89.9\,\%$ of the time over $3{,}977$ within-page pairs. The rule consults only crops already read. Asking for the page's evidence raises $p_{\mathrm{eff}}$ at the price of a smaller $c$, because the rule abstains wherever the page offers none. \textbf{Seed expansion.} With one seed per character there is nothing to average and nothing for commitment to match, so both changes stay inert. Seed expansion uses the same page groups to add, before the stream starts, the crops on a seed's page that share its character, $1.43$ per seed on average and $90.6\,\%$ of them correct.

\textbf{What the changes are worth.}\label{subsec:worth}\label{subsec:expand} Table~\ref{tab:vouch_combined} adds the changes one at a time. With five random seeds per character the cast sheet and commitment together add 3.3 to 6.5 points of identity Rank-1 over the static gallery, significant on all five backbones and 28 to 41 percent of the oracle's gain. The cast sheet alone helps every backbone and commitment adds more on four of them, while on MagiV2, the strongest gallery, the cast sheet is marginally better alone. With one seed per character, seed expansion adds 4.7 to 6.8 points on all but MagiV2. Zero-shot on Manga109 the cast sheet helps both manga-native backbones and commitment lifts MagiV3 further, while MagiV2, whose gallery starts strongest there, keeps the cast sheet alone as Section~\ref{sec:condition} predicts.
\section{Chronological seeding, and a binding built for it} \label{sec:binding} \label{subsec:seqt}

\begin{figure}[t]
\centering \includegraphics[width=\linewidth]{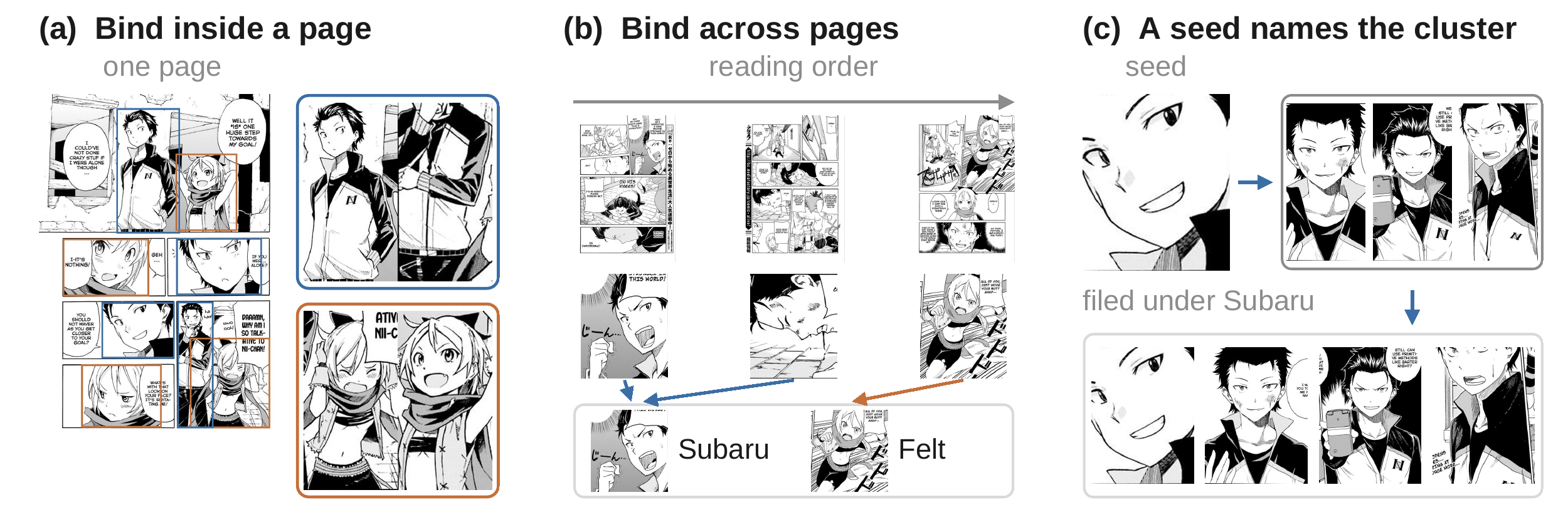}\\[3pt]
\includegraphics[width=\linewidth]{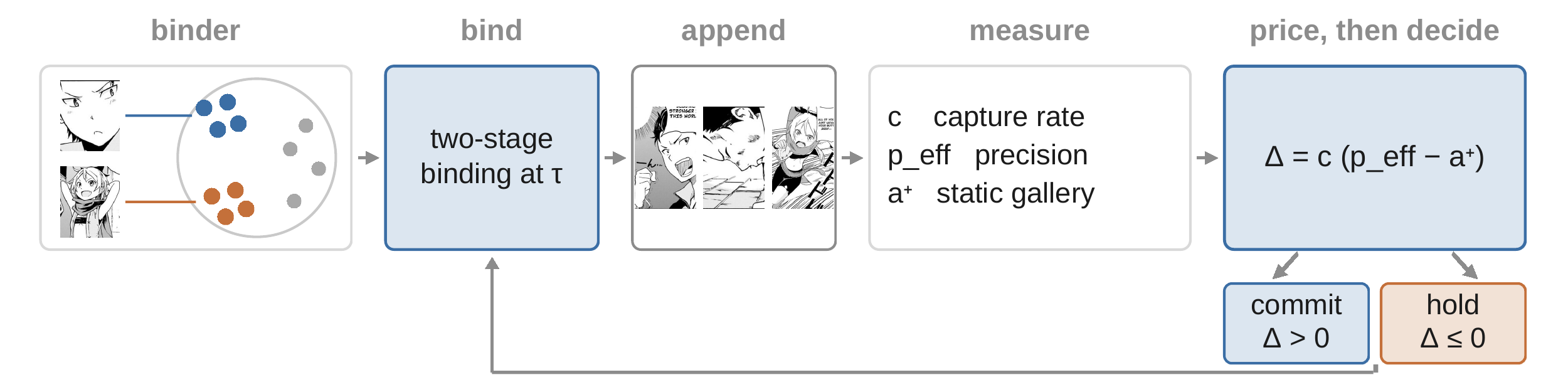}
\caption{\textbf{Binding, and pricing the binder.} The top row uses Re:Verse's Re:Zero annotations~\citep{baranwal2025reverse}. (a) A page's crops are grouped above one similarity threshold, (b) page groups merge in reading order into the best earlier match above a second or start new ones, and (c) a group's first seed names it. In the bottom row any frozen encoder binds with threshold $\tau$, and its crops are committed when Equation~\ref{eq:commit} predicts a positive $\Delta$, with $p_{\mathrm{eff}}$ the additions' precision on the queries they capture and $a^{+}$ the static gallery's accuracy there. That prediction, never the measured gain, chooses $\tau$.} \label{fig:vouch_binding}\label{fig:vouch_commit_flow}
\end{figure}

\begin{figure}[t]
\centering \includegraphics[width=0.48\linewidth]{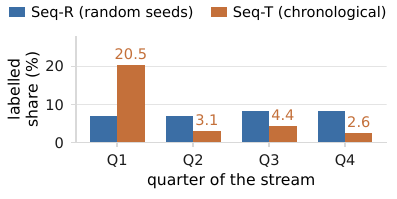} \caption{\textbf{Where the supervision sits.} Labelled share of the last twenty crops read, by quarter of the stream, over the 8 held-out \PopChars series at $k=5$. Both seedings place the same number of labels in the volume and distribute them differently.} \label{fig:supervision}
\end{figure}

\textbf{Where the supervision sits.}\label{subsec:supervision} When the seeds are each character's first appearances, the changes of Section~\ref{sec:vouch} have little to act on, and the reason is where the labels sit. In the opening quarter of the stream a fifth of the last twenty crops read are labelled, and afterwards fewer than one in twenty (Figure~\ref{fig:supervision}). A typical query sits four times further from its nearest seed than with random seeds, and that distance alone accounts for almost all of the gap between the two regimes (Appendix~\ref{app:seqt}). Commitment finds a quarter as much of the stream to act on, and seed expansion little to add, since first appearances crowd into a few opening pages. What a query needs is a nearby reference.

\textbf{One operation, applied twice.}\label{subsec:binding_op} Binding supplies that reference. A \emph{binder}, a frozen embedding for comparing crops, groups a page's crops above a similarity threshold, and page groups then merge in reading order into the best earlier match above a second threshold or open a new one (Figure~\ref{fig:vouch_binding}, top). A group's first seed names it and later members join the gallery under that name, with no detector or panel geometry involved. Our shared binder is MagiV2's fine-tuned embedding at thresholds 0.7 and 0.5 for every backbone. %
\begin{table}[t]
\centering \footnotesize \setlength{\tabcolsep}{3pt} \renewcommand{\arraystretch}{1.0}
\caption{\textbf{Two-stage binding under chronological seeding.} Change in P4 identity Rank-1 over the static gallery ($a$ on \PopChars) at $k=5$, averaged over series and seed draws, and \textbf{bold} is a gain at $p<0.05$ paired over series. Columns are \PopChars\ unless marked Manga109 (27 volumes, zero-shot). \emph{Shared} is the binder of Section~\ref{sec:binding} for every backbone and \emph{Self} each backbone's own. \emph{Oracle} is the same representation's ceiling ($+20.2$ to $+26.4$ on Manga109), \emph{Shuffled} permutes the added identities ($-8.1$ to $-43.5$ on Manga109), and \emph{Seq-R} the same binding with random seeds.}
\label{tab:transport}
\begin{tabular}{l|c|cc|c|c|cc}
\toprule
\rowcolor{gray!15}
& & \multicolumn{2}{c|}{\textbf{Shared binder}} & \textbf{Self} & & \multicolumn{2}{c}{\textbf{Controls}} \\
\rowcolor{gray!15}
\multirow{-2}{*}{\textbf{Backbone}} & \textbf{$a$} & \textbf{\PopChars} & \textbf{Manga109} & \textbf{\PopChars} & \textbf{Oracle} & \textbf{Shuffled} & \textbf{Seq-R} \\
\midrule
TransReID     & 20.7 & \textbf{+14.32} & \textbf{+10.32} & +4.96 & +20.94 & -5.56 & +3.93 \\
InstructReID  & 20.4 & \textbf{+15.45} & \textbf{+10.34} & +4.60 & +22.46 & -5.93 & +3.86 \\
ReID5o        & 20.9 & \textbf{+16.94} & \textbf{+10.35} & +2.52 & +24.63 & -6.07 & +2.70 \\
MagiV3        & 26.5 & \textbf{+16.45} & \textbf{+8.00} & +1.73 & +23.56 & -10.71 & +1.89 \\
MagiV2        & 37.4 & +12.51 & -7.44 & +8.46 & +21.25 & -20.44 & -4.92 \\
\bottomrule
\end{tabular}
\end{table}
Its additions are right about half the time on both corpora, more often than any \PopChars\ static gallery (Table~\ref{tab:transport}). With first-appearance seeds it adds 12.5 to 16.9 points of identity Rank-1 on every backbone on \PopChars, about two thirds of the oracle's gain, and 8.0 to 10.4 on four of five backbones over Manga109. The carried identity is what pays. The same crops added under shuffled identities cost points instead, and with random seeds, where a nearby reference already exists, binding adds nothing significant.

\textbf{The binder is a free choice, and Equation~\ref{eq:commit} prices it.} Each backbone can instead bind with its own embedding at a threshold chosen by the predicted $\Delta$ rather than the measured gain (Figure~\ref{fig:vouch_binding}, bottom). That helps all five backbones but less than the shared binder, because the four weaker embeddings bind barely more precisely than their galleries already answer, while MagiV2's own embedding binds at nearly twice the precision of the others (Appendix~\ref{app:seqt}).

\textbf{The commit condition predicts the one loss.}\label{subsec:ceiling} With each backbone's own binder a binding captures most queries, so $p_{\mathrm{eff}}$ stays near the binder's precision and $a^{+}$ near the average accuracy $a$ (Appendix~\ref{app:seqt}), and the condition becomes precision against $a$. Nine of ten backbone and corpus galleries start below the shared binder's precision on their corpus, and all nine gain. The tenth, MagiV2 on Manga109, starts above it at 57.5 against 47.8 and is the only one that loses. The same reading says why abstention is the mechanism. An addition that takes over queries the gallery already answers raises $a^{+}$ and not $p_{\mathrm{eff}}$, so every gain comes from a rule that chooses when to act.
\section{Conclusion, limitations, and future directions}
\label{sec:conclusion}

Readers recognise characters as a story unfolds, without knowing its full cast in advance. \Recog brings this setting to character Re-ID by replacing the fixed gallery of conventional evaluations with one that models must build and maintain as they read. This changes the task from simply matching appearances to deciding which matches deserve to become references for future recognition. Our experiments reveal a surprising divide. Recognising familiar characters is within reach, but knowing which matches to trust is not. Models that grow their galleries from their own predictions perform worse than those that never update them. The problem is not merely recognising a character, but preventing uncertain predictions from corrupting future recognition. A single mistaken update can affect decisions throughout the remaining story, making the quality of memory as important as the quality of the visual representation.

The commit condition formalises this trade-off by identifying when adding a match improves the gallery rather than weakening it. It provides a common basis for understanding and bounding the update mechanisms we study. Guided by this condition, \Vouch commits new observations only when the page itself provides sufficient evidence. Without fitting any parameters, it recovers between one-third and two-thirds of the gains available from perfect identity labels. These results show that better evidence selection can substantially improve identity maintenance without further adapting the Re-ID backbone. Binding strengthens this result by preserving first-appearance references throughout the stream. Its gains suggest that reliable identity anchors need not come from an external cast list and the story can establish them as it unfolds. Dialogue offers a natural extension, where character names could connect visual appearances to persistent identities and help resolve ambiguous matches.

Together, \Recog and \Vouch separate two challenges that conventional Re-ID evaluations often conflate, recognising an identity and maintaining trustworthy evidence of it. \Recog measures the gap, while \Vouch shows how to narrow it. Testing the commit condition wherever a system grows its own reference set, as in tracking and self-training, is a natural next step. Reliable recognition requires more than finding the right match; it requires knowing when to remember it.

\noindent\textbf{Limitations.}\label{sec:limitations}
Our claims concern identity maintenance. The P3 threshold is a measurement choice, and supervised metric learning may produce the compactness P3 rewards without improving identity structure. Our corpora do not yet test appearance changes across long story arcs. All experiments use Japanese comics and ground-truth crops, leaving Western comics and webtoons for future work.

\FloatBarrier

\begin{ack}
This work received no external funding. The authors declare no competing interests.
\end{ack}

\bibliographystyle{plainnat}
\bibliography{main}

\clearpage
\appendix
\raggedbottom
\setcounter{topnumber}{3}
\setcounter{bottomnumber}{2}
\setcounter{totalnumber}{5}
\renewcommand{\topfraction}{0.92}
\renewcommand{\bottomfraction}{0.85}
\renewcommand{\textfraction}{0.06}
\renewcommand{\floatpagefraction}{0.72}
\setlength{\textfloatsep}{6pt plus 2pt minus 2pt}
\setlength{\floatsep}{5pt plus 2pt minus 2pt}
\setlength{\intextsep}{5pt plus 2pt minus 2pt}
\suppressfloats[t]

\begin{center}
\Large\textbf{\Recog \ - A Framework For Open-Set \\Comic Character Re-Identification}\\[0.5em]
\large (\textit{Supplementary Material})
\end{center}

\section{Corpora, protocols and evaluation} \label{app:impl_details}

\setlength{\textfloatsep}{6pt plus 2pt minus 2pt} \setlength{\floatsep}{6pt plus 2pt minus 2pt} \setlength{\intextsep}{6pt plus 2pt minus 2pt} \setlength{\dblfloatsep}{6pt plus 2pt minus 2pt} \setlength{\dbltextfloatsep}{6pt plus 2pt minus 2pt} \setlength{\abovecaptionskip}{4pt} \setlength{\belowcaptionskip}{0pt} \renewcommand{\floatpagefraction}{0.8}

\subsection{Evaluation corpora} \label{app:datasets}

\textbf{\PopChars}~\citep{sachdeva2024tails}. The primary in-domain evaluation corpus, derived from the publicly released character-name annotations of the PopManga corpus and providing crop-level identities across 23 contemporary manga series. We use the series-disjoint split distributed with our framework code, with 13 training series carrying 198 characters and 7{,}668 crops, 2 development series (Dragon Ball and Kuroko's Basketball) carrying 10 characters and 873 crops, and 8 held-out test series carrying 70 characters and 4{,}058 crops, with no character, page, or panel overlap between splits. The test series are Bakuman, Demon Slayer, Dr Stone, Hunter $\times$ Hunter, Kagurabachi, Nisekoi, Oshi No Ko, and Tokyo Ghoul (Table~\ref{tab:suppl_per_manga} reports per-manga numbers). Crops are unique instances. A character that appears several times on one page contributes several crops, while cross-page recurrence is the unit on which P2 and P4 are scored. The corpus is released for research use only and the underlying raw page images remain with their original publishers.

\textbf{Manga109}~\citep{Aizawa_2020}. Used for cross-corpus transfer and only zero-shot, so no model is trained on it. We evaluate on the 27 volumes that the volume-disjoint 82/27 split released with our code holds out, which carry 784 characters and 29{,}315 crops. The XML annotations carry the crop-level character-identity structure the protocols require, and we convert them to the same series-folder layout used for \PopChars via \texttt{scripts/convert\_manga109.py}. Manga109 character labels are volume-internal rather than series-wide, so the evaluation in Appendix~\ref{app:cross_dataset} scores each volume on its own and macro-averages over the 27. Manga109 is distributed under a research-only academic license that requires institutional registration. We redistribute the conversion script and not the images.

\textbf{Re:Verse}~\citep{baranwal2025reverse}. A small cross-page consistency benchmark on the Re:Zero manga, scored with the same protocol code. The released character crops cover 12 main and recurring characters across 1{,}825 instances drawn from 308 pages of the Re:Zero anthology. We evaluate on the closed-set retrieval split (P1) only, since Re:Verse was designed primarily as a probe of cross-page identity consistency rather than full streaming evaluation. The Re:Verse comparison numbers in Table~\ref{tab:reverse_results} use the same backbone checkpoints reported elsewhere in the paper, so the gap between vision-language models and specialist Re-ID encoders reflects representation quality rather than fine-tuning.

\begin{figure}[!ht]
\centering \includegraphics[width=\columnwidth]{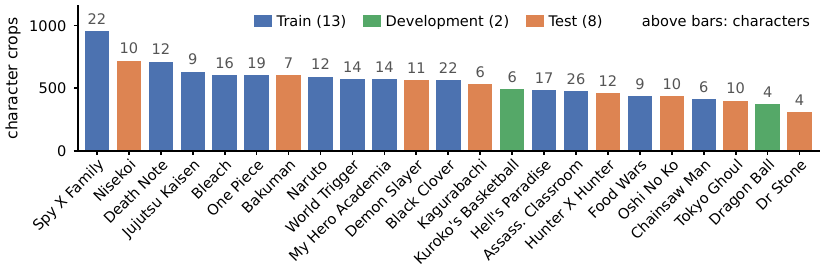} \caption{\textbf{Crops per series} in \PopChars, coloured by split into 13 training series, 2 development series and the 8 test series. Character counts above bars.} \label{fig:suppl_crops_per_series}
\end{figure}

\begin{figure}[!ht]
\centering \includegraphics[width=\columnwidth]{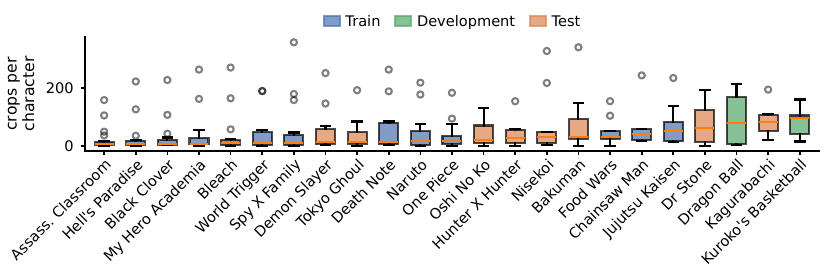} \caption{\textbf{Crop count distribution per series.} Series with large casts (left) exhibit heavier tails in per-character crop counts.} \label{fig:suppl_boxplot}
\end{figure}

\paragraph{Per-series statistics.}\label{app:per_series} Table~\ref{tab:datasets} gives the split sizes of the three corpora. \PopChars and Manga109 both have heavy-tailed character-frequency distributions, and the per-series shape is reported here so the macro-averaged numbers in the body are interpretable. Figure~\ref{fig:suppl_crops_per_series} gives the crop count of every \PopChars series and the split it belongs to. Figure~\ref{fig:suppl_boxplot} gives the per-character crop counts of each series and exposes the heavy tail. A handful of characters per series carry most of the crops, while the secondary cast appears in only a few panels. Every metric averages over the queries of a series and then over series, so each series weighs the same while, within a series, the frequent characters supply most of the queries. P2 gives every character the same $k$ seeds, so its gallery favours no one.

\begin{table}[H]
\centering \footnotesize \setlength{\tabcolsep}{4pt} \renewcommand{\arraystretch}{1.02} \caption{\textbf{Dataset statistics.} Splits are series-disjoint, and volume-disjoint on Manga109. Avg C/Ch is the average crops per character. No model is trained on Manga109 or Re:Verse, so every number on them is zero-shot.} \label{tab:datasets}
\begin{tabular}{l|l|c|c|c|c}
\toprule
\rowcolor{gray!15}
\textbf{Dataset} & \textbf{Split} & \textbf{\#Series} & \textbf{\#Chars} & \textbf{\#Crops} & \textbf{Avg C/Ch} \\
\midrule
\multirow{3}{*}{POPCharacters}
  & Train       & 13 & 198 & 7{,}668 & 38.7 \\
  & Development &  2 &  10 &   873 & 87.3 \\
  & Test        &  8 &  70 & 4{,}058 & 58.0 \\
\midrule
Manga109        & Test  & 27 &   784 & 29{,}315 & 37.4 \\
\midrule
Re:Verse        & Test  &  1 &   12 & 1{,}825 & 152.1 \\
\bottomrule
\end{tabular}
\end{table}

\subsection{Protocols and evaluation harness} \label{app:reeval}

\paragraph{Protocol implementation.} P1 puts $\max(1,\lfloor 0.2\,n\rfloor)$ random crops of a character with $n$ crops in the gallery and queries with the rest. Characters with a single crop are excluded and counted. P2 draws $k$ seeds per character, uniformly at random (Seq-R) or the first $k$ in reading order (Seq-T), and a character with at most $k$ crops enters the gallery only. P3 streams every crop of each held-out series in reading order and compares each crop with the normalised running centroid of every cluster. A crop above $\tau_{\mathrm{nov}}\!=\!0.55$ joins the most similar cluster and otherwise opens a new one. P4 starts from the P2 gallery and answers the queries in reading order. Each query is scored against the gallery as it stood before the query arrived and is then appended to its top-1 identity's buffer, of capacity $B_{\max}\!=\!50$, with the seeds protected from eviction. All four protocols operate on ground-truth detections rather than end-to-end localisation, which isolates Re-ID from detection.

\paragraph{Harness decisions.} Every number in this paper comes from one harness, and four of its decisions move absolute values.

\textbf{Reading order is the dataset's.} Page order comes from each corpus's own page numbering, with the interpreter's hash seed pinned so that a run reproduces. Deriving it from a filename pattern instead falls back to a per-process string hash on these corpora, which makes a Seq-T stream non-chronological and stops any two P2 or P4 runs from being comparable.

\textbf{The reported checkpoint is the last epoch.} Every model is read at its final epoch, and the development split enters no table. Selecting a checkpoint on a ten-identity development split whose own spread is about three points stops runs early and selects on noise.

\textbf{MagiV2's masking is disabled at inference.} Its crop encoder is a masked autoencoder whose patch masking stays active in evaluation mode unless it is switched off, as the authors' own interface does. Left on, every MagiV2 number is a random draw.

\textbf{P4 is scored on identity Rank-1.} Exemplar-level average precision with one relevant gallery entry per query is $1/\mathrm{rank}$ and therefore moves with the size of the gallery, which is precisely the quantity gallery growth changes, so exemplar mAP cannot compare a static gallery with a grown one. Identity Rank-1 asks what the protocol is about, whether the system names the right character.

\paragraph{Reproducibility.} The split files, \texttt{data\_split.yaml} for the \PopChars training, development and test series and \texttt{data\_split\_manga109.yaml} for the Manga109 volumes, are released with the framework code and pin the exact series memberships. Every metric reported in this paper is computed by the released evaluation harness and is reproducible from the (backbone, configuration, training run, protocol, seed) tuple, with per-tuple JSON results released alongside the code. Galleries are drawn with seeds $\{0,\ldots,4\}$ on \PopChars and Re:Verse and $\{0,1,2\}$ on Manga109. Every trained configuration is trained with seeds 0, 1 and 2, and Manga109 and Re:Verse are scored with the first of those runs.

\paragraph{Measurement validity.}\label{sec:validity} A framework is worth only as much as the differences it can resolve, and three quantities bound every claim here, each measured rather than assumed.

\textbf{What a difference has to clear.} Every headline cell is trained three times, and the spread between those runs is the floor a difference has to clear before it means anything. Measured over 28 configurations with three complete seeds each, the median standard deviation of a \MeCha run is 0.08 P1 mAP, 0.30 P1 Rank-1, 0.15 P2-R@1 mAP, 0.47 P2-T@1 mAP, 0.51 P4-R@1 identity Rank-1 and 1.36 P4-T@1 identity Rank-1. Without memory the same figures are three to seven times smaller. Two consequences run through the paper. Rank-1 is roughly four times noisier than mAP, so a claim about naming a character needs four times the margin of a claim about ranking. And P4 under chronological seeding is the noisiest cell the framework has, which is why the \Vouch\ comparisons in Section~\ref{sec:vouch} are paired over series and reported with a test rather than as a difference of two means.

\textbf{Deciding growth on a new corpus.} Section~\ref{sec:condition} decides growth from terms measured on a labelled slice of the corpus at hand. The test labels a random half of a corpus's held-out series or volumes, measures $c$, $p_{\mathrm{eff}}$ and $a^{+}$ there, and decides growth for the other half. Over 200 random splits the call is correct every time on six of the seven cells of Table~\ref{tab:commit}, including MagiV2 on Manga109, where the intuitive test says grow and growth loses 6.23. The seventh, MagiV2 on \PopChars, has a true effect of $-0.04$, whose sign carries no information. The decision travels but the magnitude does not, since \PopChars\ terms applied to the Manga109 MagiV2 cell predict $+0.33$ against a measured $-6.23$, which is why the terms are measured on the corpus the decision is for.

\textbf{Chance levels.} A random ranking scores approximately 33 mAP at both P1 and P2. Chance Rank-1 is 30.1 at P1 against 12.9 at P2 at $k=1$, because the P1 gallery holds many entries per identity and the P2 gallery holds one. That difference is what lets a single seed exceed the P1 mAP ceiling while recovering under two thirds of its Rank-1 (Section~\ref{subsec:ceiling_and_growth}).

\subsection{Backbones and configurations} \label{app:backbones}

Table~\ref{tab:backbones} lists the five backbones, their pre-training regimes, input resolutions, and feature dimensions. Each keeps its native width $d$, at which its BNNeck and memory block operate, so MagiV3 runs at 1024 dimensions and ReID5o at 512, with no adapter to a shared space.

\begin{table}[t]
\centering \footnotesize \setlength{\tabcolsep}{4pt} \renewcommand{\arraystretch}{1.02} \caption{\textbf{Backbone architectures} evaluated by \Recog. $d$ is the native feature width, at which each backbone's BNNeck and memory block operate.} \label{tab:backbones}
\resizebox{\linewidth}{!}{%
\begin{tabular}{l|l|l|l|c}
\toprule
\rowcolor{gray!15}
\textbf{Backbone} & \textbf{Architecture} & \textbf{Pre-training} & \textbf{Input} & \textbf{$d$} \\
\midrule
TransReID~\citep{he2021transreid} & ViT-B/16 & ImageNet + Re-ID & $256\!\times\!128$ & 768 \\
MagiV2 & ViT-B & Manga embeddings & $224\!\times\!224$ & 768 \\
MagiV3 & Florence-2~\citep{yuan2021florence} & Manga comprehension & $384\!\times\!384$ & 1024 \\
InstructReID~\citep{he2023instructreid} & ViT-B/16 & Multi-modal Re-ID & $256\!\times\!128$ & 768 \\
ReID5o & CLIP ViT-B/16~\citep{radford2021clip} & CLIP + Re-ID & $384\!\times\!128$ & 512 \\
\bottomrule
\end{tabular}%
}%
\end{table}

\paragraph{Configuration matrix.}\label{app:configs} Every backbone in the main paper is evaluated under five configurations, summarised in Table~\ref{tab:suppl_configs}. The matrix varies three design axes, whether a BNNeck is trained on the backbone (rows 2--5), whether \MeCha is enabled (rows 3 and 5) and whether LoRA adapts the backbone (rows 4 and 5). The \emph{Pretrained} row is the released backbone scored as it is, its class token L2-normalised and matched directly with no BNNeck and no memory. The other four rows train a BNNeck on \PopChars, with and without \MeCha and LoRA. This factorisation lets the full P1 grid in Appendix~\ref{app:full_grid} attribute gains to fine-tuning, memory and LoRA in turn.

\begin{table}[!ht]
\centering \footnotesize \setlength{\tabcolsep}{5pt} \renewcommand{\arraystretch}{1.1} \caption{Five configurations analysed per backbone, with BNNeck fine-tuning, \MeCha and LoRA toggled along three axes.} \label{tab:suppl_configs}
\begin{tabular}{l|c|c|c|c}
\toprule
\rowcolor{gray!15}
\textbf{Configuration} & \textbf{Backbone} & \textbf{BNNeck} & \textbf{Memory} & \textbf{LoRA} \\
\midrule
Pretrained                & frozen  & --      & --        & -- \\
Finetuned                 & frozen  & trained & --        & -- \\
Finetuned + Memory        & frozen  & trained & trained   & -- \\
Finetuned + LoRA          & adapted & trained & --       & trained \\
Finetuned + Memory + LoRA & adapted & trained & trained   & trained \\
\bottomrule
\end{tabular}
\end{table}

\FloatBarrier
\section{The \MeCha baseline} \label{app:mecha}

\subsection{Architecture} \label{app:mecha_arch}

\begin{figure}[!ht]
\centering \includegraphics[width=\linewidth]{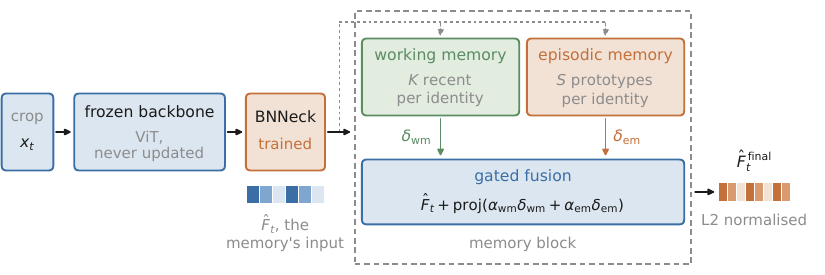} \caption{\textbf{\MeCha.} A frozen backbone produces L2-normalised features via BNNeck. Working Memory (cross-attention over recent observations) and Episodic Memory (cross-attention over identity prototypes) produce residuals $\delta^{wm}_t,\delta^{em}_t$, combined by Gated Fusion through a small-init projection. Optional LoRA adapts the last $L$ backbone attention layers. The MLPs and attention projections are trained, the backbone $\mathcal{B}$ is frozen, and the prototypes $\mathbf{P}_c$ and FIFO buffers $\mathbf{B}_c$ are runtime state.} \label{fig:mecha_arch_appendix}
\end{figure}

A frozen backbone produces $F_t=\mathcal{B}(x_t)$ for a crop $x_t$, and a BatchNorm neck (BNNeck) L2-normalises it to $\hat{F}_t$, which keeps cosine similarity stable between training and inference and separates the learnable memory parameters from the memory contents. Two branches then read memory (Figure~\ref{fig:mecha_arch_appendix}). Working memory attends over each character's most recent features, and episodic memory attends over a bank of prototypes that tracks distinct appearance modes. Each branch emits a pure residual, and a gated fusion adds them through one small-initialised projection, so memory starts as a near no-op and earns its contribution through training. At $d=768$ the block adds 10.1M trainable parameters (Appendix~\ref{app:cost}).

\textbf{Notation.} $K$ is the working-memory capacity per character and $S$ the number of episodic prototype slots. $P$ and $K_\text{supp}$ are the identities per PK batch and the support crops per identity, $\rho$ is the ID-drop probability, $\tau$ the prototype-loss temperature and $\tau_{\mathrm{nov}}$ the P3 novelty threshold. $\mathbf{B}_c$ is the FIFO buffer and $\mathbf{P}_c$ the prototype bank of character $c$, $\hat{F}_t$ is the BN-normalised backbone feature and $\hat{F}_t^{\,\mathrm{final}}$ the memory-enhanced one, and $\hat{c}_t$ is the identity predicted in Pass~1. The two residuals are $\delta^{wm}_t$ and $\delta^{em}_t$, and the fusion weights $\alpha^{wm}$ and $\alpha^{em}$.

\textbf{Working Memory.} For each character $c$ the FIFO ring buffer $\mathbf{B}_c$ keeps the $K$ most recent features, and multi-head cross-attention~\citep{vaswani2017attention} reads it as
\begin{equation}
\delta_t^{wm} = g_\phi\!\left(\mathrm{LN}\!\left(\mathrm{MHCA}\!\left(\mathrm{LN}(\hat{F}_t)W_Q,\;\mathrm{LN}(\mathbf{B}_c)W_K,\;\mathrm{LN}(\mathbf{B}_c)W_V\right)\right)\right)
\label{eq:wm}
\end{equation}
with learned projections $W_Q,W_K,W_V\in\mathbb{R}^{d\times d}$, LayerNorm~\citep{ba2016layer} and $g_\phi$ a two-layer MLP (GELU, Kaiming init). The residual is zero while the buffer is empty.

\textbf{Episodic Memory.} A non-parametric prototype bank $\mathbf{P}_c=\{\mu_c^1,\ldots,\mu_c^S\}$ per character is initialised from a support set by farthest-point sampling~\citep{qi2017pointnet}. A new observation $f_t$ replaces the most similar slot, so the bank tracks new appearance modes instead of blurring them as an exponential average would,
\begin{equation}
j^* = \arg\max_{j\in\{1,\ldots,S\}}\cos(f_t,\mu_c^j),\qquad \mu_c^{j^*}\leftarrow f_t.
\label{eq:proto_update}
\end{equation}
The branch queries $\mathbf{P}_c$ in identity-guided mode or the union $\mathbf{P}=\bigcup_{c'}\mathbf{P}_{c'}$ in search-all mode,
\begin{equation}
\delta_t^{em} = g_\psi\!\left(\mathrm{LN}\!\left(\mathrm{MHCA}\!\left(\mathrm{LN}(\hat{F}_t),\;\mathrm{LN}(\mathbf{P}),\;\mathrm{LN}(\mathbf{P})\right)\right)\right).
\label{eq:em}
\end{equation}

\textbf{Gated Fusion.} A learned gate over $[\hat{F}_t;\delta_t^{wm};\delta_t^{em}]$ produces the final descriptor,
\begin{align}
[\alpha^{wm},\alpha^{em}] &= \mathrm{Softmax}\!\left(f_\gamma([\hat{F}_t;\delta_t^{wm};\delta_t^{em}])\right),\label{eq:gate_app}\\
\hat{F}_t^{\,\mathrm{final}} &= \frac{\hat{F}_t + g_\omega\!\left(\alpha^{wm}\delta_t^{wm}+\alpha^{em}\delta_t^{em}\right)}{\bigl\|\hat{F}_t + g_\omega\!\left(\alpha^{wm}\delta_t^{wm}+\alpha^{em}\delta_t^{em}\right)\bigr\|_2}.
\label{eq:fusion_app}
\end{align}
Here $f_\gamma$ is a two-layer MLP ($\mathbb{R}^{3d}\!\to\!\mathbb{R}^d\!\to\!\mathbb{R}^2$) with LayerNorm and GELU, and $g_\omega$ is a two-layer MLP whose final layer is initialised at $\mathcal{N}(0,0.001^2)$. Equation~\ref{eq:fusion_app} is the only residual path.

\paragraph{Why the ablation removes episodic memory.}\label{app:fps}\label{app:episodic_training} The prototype bank holds $S=5$ slots per character, initialised by farthest-point sampling over the support set so the slots span the character's appearance manifold instead of clustering on its mode. Training pairs each query crop with support crops drawn from earlier pages of the same chapter, so the bank a query meets in training resembles the one it meets at test time, and identity dropout at $\rho=0.5$ forces the search-all path that the open-set protocols use. Prototypes stay in clean BN-normalised space and are never re-extracted through memory, which avoids a circular dependency between the bank and the block that reads it.

The ablation of Appendix~\ref{app:ablation} removes this branch and finds that working memory carries the whole effect of the block, which is why \Vouch{} puts its memory in the gallery and not in the encoder.

\subsection{Training, inference and cost} \label{app:training_details}

\textbf{Reading-order-aware episodic training.} Each mini-batch uses PK sampling ($P$ characters, $K$ instances), split into support ($K_\text{supp}$) and query sets, and memory is reset at every step. The support set initialises episodic memory by farthest-point sampling and primes the working-memory buffers. Pass~1 runs under \texttt{torch.no\_grad()} in search-all mode, and the gradient-enabled Pass~2 routes working memory by $\hat{c}_t$. ID-drop silences the episodic identity signal with probability $\rho$ so that the gate cannot come to depend on it,
\begin{equation}
c_t^{\mathrm{EM}} = \begin{cases}\varnothing & \text{w.p. }\rho\\ \hat{c}_t & \text{otherwise}\end{cases},\qquad
c_t^{\mathrm{WM}} = \hat{c}_t \text{ (always)}.
\label{eq:id_drop_app}
\end{equation}

\textbf{Loss.} The combined objective is $\mathcal{L} = \lambda_\text{proto}\mathcal{L}_\text{proto} + \lambda_\text{trip}\mathcal{L}_\text{trip} + \lambda_\text{mem}\mathcal{L}_\text{mem} + \lambda_\text{CE}\mathcal{L}_\text{CE}$. Its terms are a prototype loss (cosine cross-entropy with $\tau\!=\!0.15$), a batch-hard triplet loss~\citep{hermans2017defensetripletlossperson} (margin $0.3$, cosine distance), an InfoNCE memory-consistency loss~\citep{oord2018representation,chen2020simclr} that aligns memory-enhanced and pre-memory features, and an auxiliary cross-entropy. The weights are $\lambda_\text{proto}=\lambda_\text{trip}=1.0$ and $\lambda_\text{mem}=0.1$, and $\lambda_\text{CE}$ is $0.3$ without \MeCha and $0$ with it, since \MeCha discards its classifier.

\textbf{Hyperparameters.} $K\!=\!8$ working-memory slots, $S\!=\!5$ episodic prototypes, 8-head cross-attention, dropout 0.1 and ID-drop $\rho\!=\!0.5$. Training runs for 200 epochs with AdamW~\citep{loshchilov2018adamw} at lr $10^{-4}$ and weight decay $10^{-4}$, a 5-epoch warmup with cosine annealing~\citep{loshchilov2017sgdr} and FP16 mixed precision. PK sampling uses $P\!=\!8$ and $K\!=\!4$ ($K_\text{supp}\!=\!K_\text{query}\!=\!2$), with $P\!=\!4$ for MagiV3, whose $384\!\times\!384$ inputs need more memory. LoRA uses $r\!=\!8$ and $\alpha\!=\!16$ on the last $L\!=\!4$ attention layers at lr $10^{-5}$.

\textbf{Compute.} Each training run takes 0.45 to 1.27 GPU-hours for 200 epochs on one Turing- or Ampere-class GPU, so the 84 runs behind the reported numbers take about 60 GPU-hours, from four trained configurations of five backbones and four ablations of two backbones at three training runs each. Evaluating one checkpoint on every protocol takes minutes per series on one GPU, and the full evaluation, dominated by the 27 Manga109 volumes, takes a few tens of GPU-hours.

\begin{figure}[!ht]
\centering \includegraphics[width=\linewidth]{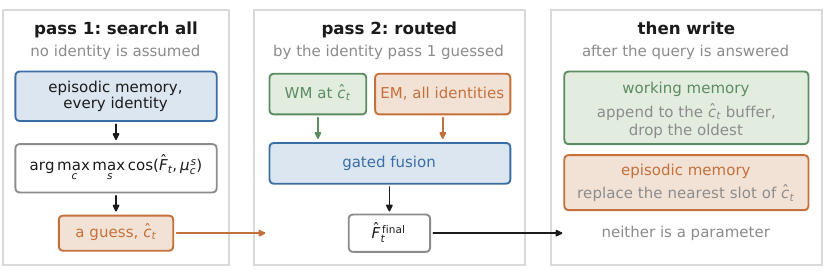} \caption{\textbf{Two-pass open-set inference} schematic. Pass~1 produces a coarse identity hypothesis. Pass~2 uses the hypothesis to route Working Memory and emits the final descriptor. The FIFO buffer and prototype bank update only after Pass~2.} \label{fig:suppl_two_pass_protocol}
\end{figure}

\paragraph{Two-pass open-set inference.}\label{app:two_pass} At test time \MeCha reads the stream one crop at a time with no gradient and routes each crop through the memory block twice (Figure~\ref{fig:suppl_two_pass_protocol}, Algorithm~\ref{alg:two_pass}). A search-all pass produces a coarse identity guess, and a refined pass uses that guess to pick the working-memory buffer, since the buffer is chosen by predicted identity and Pass~1 is what predicts it. The refined descriptor then updates the buffer and the prototype bank. This one procedure serves all four protocols without retraining, and only the composition of the initial gallery $\mathcal{G}_0$ changes from P1 to P4.

\begin{algorithm}[!ht]
\caption{Two-Pass Open-Set Inference}
\label{alg:two_pass}
\textbf{Input:} stream $\{x_t\}_{t=1}^T$, frozen $\mathcal{B}$, trained memory block $\mathcal{M}$ and initial gallery $\mathcal{G}_0$. \quad
\textbf{Output:} predictions $\{\hat c_t\}_{t=1}^T$ and descriptors $\{\hat F_t^{\,\mathrm{final}}\}_{t=1}^T$.
\begin{algorithmic}[1]
\STATE Initialise EM prototypes from $\mathcal{G}_0$ by farthest-point sampling and prime the WM buffers from $\mathcal{G}_0$.
\FOR{$t = 1, \ldots, T$}
  \STATE $\hat F_t \leftarrow \mathrm{BNNeck}(\mathcal{B}(x_t))$ \hfill\COMMENT{backbone feature}
  \STATE \textit{Pass 1 (search-all):}
  \STATE \quad $\delta_t^{em,(1)} \leftarrow \mathrm{EM.query}(\hat F_t,\, c=\varnothing)$
  \STATE \quad $\hat F_t^{(1)} \leftarrow \mathrm{Fuse}(\hat F_t,\, \mathbf{0},\, \delta_t^{em,(1)})$
  \STATE \quad $\hat c_t \leftarrow \arg\max_{c,\,s}\, \cos(\hat F_t^{(1)},\, \mu_c^s)$ \hfill\COMMENT{coarse identity guess}
  \STATE \textit{Pass 2 (predicted routing):}
  \STATE \quad $\delta_t^{wm} \leftarrow \mathrm{WM.query}(\hat F_t,\, c=\hat c_t)$
  \STATE \quad $\delta_t^{em,(2)} \leftarrow \mathrm{EM.query}(\hat F_t,\, c=\varnothing)$ \hfill\COMMENT{still search-all}
  \STATE \quad $\hat F_t^{\,\mathrm{final}} \leftarrow \mathrm{Fuse}(\hat F_t,\, \delta_t^{wm},\, \delta_t^{em,(2)})$
  \STATE \textit{Memory updates:}
  \STATE \quad $\mathrm{WM.update}(\hat F_t,\, \hat c_t)$ \hfill\COMMENT{FIFO into the predicted character's buffer}
  \STATE \quad $\mathrm{EM.update}(\hat F_t^{\,\mathrm{final}},\, \hat c_t)$ \hfill\COMMENT{replace most-similar prototype}
\ENDFOR
\end{algorithmic}
\end{algorithm}

\paragraph{Computational cost.}\label{app:cost} \MeCha is a fixed cost on top of the backbone (Table~\ref{tab:suppl_cost}). It adds 10.1M trainable parameters at $d=768$, 17.9M at MagiV3's 1024 and 4.5M at ReID5o's 512, which is 12\,\% of each ViT-B backbone, 5\,\% of MagiV3's vision encoder and 6\,\% of ReID5o's. Its working-memory buffers and prototypes are filled at test time and take about 40\,KB per character at $d=768$. The block's own two passes cost about 5\,ms per crop at batch size one on every backbone, whatever its width. The two-pass inference as implemented runs the backbone once per pass, so the whole memory path costs 2.5 to 3.1 times the backbone alone. Computing the backbone feature once and reusing it in both passes would cost one backbone forward plus the block-alone column.

\begin{table}[!ht]
\centering \footnotesize \setlength{\tabcolsep}{5pt} \renewcommand{\arraystretch}{1.1} \caption{\textbf{Computational overhead} on one NVIDIA~RTX A6000 at batch size one in FP32, median over 300 crops of one test series. \emph{$+$ Memory} is the two-pass inference as implemented, which runs the backbone once per pass, and \emph{block alone} is the same two passes over cached backbone features. Backbone parameters count what the crop forward touches, since MagiV3 and ReID5o ship modules it never calls. The block's buffers and prototypes are filled at test time and are not parameters.} \label{tab:suppl_cost}
\begin{tabular}{l|c|c|c|c|c}
\toprule
\rowcolor{gray!15}
& \multicolumn{3}{c|}{\textbf{Latency (ms/crop)}} & \multicolumn{2}{c}{\textbf{Parameters}} \\
\rowcolor{gray!15}
\multirow{-2}{*}{\textbf{Backbone}} & \textbf{Backbone} & \textbf{$+$ Memory} & \textbf{Block alone} & \textbf{Backbone} & \textbf{Memory block} \\
\midrule
TransReID    & 5.8 & 17.8 & 4.9 & 85.6M & 10.1M \\
MagiV2       & 6.8 & 20.2 & 5.0 & 85.8M & 10.1M \\
MagiV3       & 24.2 & 59.6 & 4.8 & 360.7M & 17.9M \\
InstructReID & 7.2 & 20.3 & 4.9 & 85.8M & 10.1M \\
ReID5o       & 13.4 & 34.3 & 4.8 & 79.1M & 4.5M \\
\bottomrule
\end{tabular}
\end{table}

\subsection{Ablation and memory capacity} \label{app:ablation}

Table~\ref{tab:ablations} removes one component of \MeCha at a time on the two backbones where the block earns most, over the same three training runs as its parent, so every difference is paired over the 24 (series, training run) cells rather than read off two means. One component carries the decomposition. Removing working memory costs 0.59 P1 mAP on MagiV2 and 0.98 on MagiV3, which is the whole of what the block is worth over the finetuned baseline (0.54 and 1.01), while removing episodic memory, ID-drop or the memory-consistency loss moves P1 mAP \emph{upward} in all six cells, ID-drop included, since it exists only to teach episodic memory the search-all mode it uses at inference.

\begin{table}[tb]
\centering
\footnotesize
\setlength{\tabcolsep}{4pt}
\renewcommand{\arraystretch}{1.02}
\caption{\textbf{Per-component ablation of the memory block} on the two backbones where it earns most, macro over the 8 held-out \PopChars test series over three training runs. Each row removes one component from the full memory block, and no row carries LoRA. $\Delta$ is against the full memory block, paired over the 24 (series, training run) cells, and $^{*}$ marks $p<0.05$ under a two-sided exact sign test on those pairs. Working memory carries the whole effect. Removing it returns P1 mAP to the finetuned baseline on both backbones, while removing episodic memory, ID-drop or the memory-consistency loss moves P1 mAP \emph{upward}.}
\label{tab:ablations}
\begin{tabular}{l|l|ccc|c}
\toprule
\rowcolor{gray!15}
\textbf{Backbone} & \textbf{Configuration} & \textbf{P1 mAP} & \textbf{P1 R-1} & \textbf{P2-R@1 mAP} & \textbf{$\Delta$ P1 mAP vs full} \\
\midrule
MagiV2 & Finetuned (no memory) & 51.20 & 57.17 & 54.44 & $-0.54^{*}$ \\
 & Full memory block & 51.75 & 56.29 & 54.58 & -- \\
 & \quad $-$ working memory & 51.15 & 57.28 & 54.53 & $-0.59^{*}$ \\
 & \quad $-$ episodic memory & 52.00 & 56.26 & 54.62 & $+0.26$ \\
 & \quad $-$ ID-drop & 51.91 & 56.25 & 54.62 & $+0.16$ \\
 & \quad $-$ memory-consistency loss & 51.92 & 56.60 & 54.64 & $+0.18^{*}$ \\
\midrule
MagiV3 & Finetuned (no memory) & 41.35 & 48.71 & 43.65 & $-1.01^{*}$ \\
 & Full memory block & 42.35 & 47.59 & 44.44 & -- \\
 & \quad $-$ working memory & 41.38 & 48.60 & 43.67 & $-0.98^{*}$ \\
 & \quad $-$ episodic memory & 42.42 & 47.54 & 44.39 & $+0.06$ \\
 & \quad $-$ ID-drop & 42.49 & 48.03 & 44.22 & $+0.13$ \\
 & \quad $-$ memory-consistency loss & 42.50 & 47.85 & 44.82 & $+0.15$ \\
\bottomrule
\end{tabular}
\end{table}

\textbf{Memory capacity.} The Working Memory buffer size $K\!=\!8$ is set to retain roughly one page of context per character, which is the temporal scope at which character-specific recurrence becomes informative without saturating the FIFO with stale observations. The Episodic Memory slot count $S\!=\!5$ matches the FPS prototype count that maximises diversity for characters with two to four distinct appearance modes. Smaller $S$ collapses prototypes onto the dominant mode, while larger $S$ admits redundant slots once the appearance manifold is saturated. FIFO eviction (WM) and most-similar replacement (EM) keep buffer quality high at moderate capacity. We regard $K=8$ and $S=5$ as representative of the backbones evaluated here rather than as a universal optimum. They follow from the mechanism, not from a grid search, and the ablation shows that $S$ cannot carry the result, since removing episodic memory altogether raises P1 mAP on both backbones the ablation covers (Table~\ref{tab:ablations}).

\FloatBarrier
\section{Further results} \label{app:results}

\subsection{Retrieval in full (P1 and P2)} \label{app:full_grid}

Table~\ref{tab:suppl_p1_full} reports the full P1 closed-set grid that Table~\ref{tab:cross_protocol} summarises through best-row picks, and three patterns stand out. Every adaptation is small against the 33.15 mAP a random ranking of the same gallery scores. Training a BNNeck on a frozen backbone is worth 0.3 to 1.1 mAP over the released weights, and \MeCha adds 0.07 to 1.01 on top, most on the two comic-native backbones and least on TransReID. \MeCha also trades Rank-1 for mAP on all five backbones, giving up 0.9 to 1.2 Rank-1 wherever it gains mAP, so on four of the five the best mAP cell is not the best Rank-1 cell, ReID5o being the exception. LoRA is asymmetric and non-additive. On its own it is worth $+0.31$, $+0.64$ and $+0.40$ mAP on TransReID, InstructReID and ReID5o and $-0.51$ on MagiV2, while combined with \MeCha it gives the best cell on three of five backbones, MagiV2 and MagiV3 preferring \MeCha without it. The result space is therefore a backbone-conditional selection rather than a single ordering. Figure~\ref{fig:adaptation} summarises the part that matters, with BNNeck fine-tuning and \MeCha each adding little and the totals largest on the backbones weakest in domain.

\begin{table}[!ht]
\centering
\footnotesize
\setlength{\tabcolsep}{5pt}
\renewcommand{\arraystretch}{1.1}
\caption{Full P1 closed-set retrieval on the 8 held-out \PopChars series (Appendix~\ref{app:reeval}). Best per backbone in \textbf{bold}, second-best \underline{underlined}. \emph{Finetuned} trains a BNNeck on a frozen backbone, and only LoRA rows adapt backbone weights. Every row is the mean over three training runs.}
\label{tab:suppl_p1_full}
\begin{tabular}{l|l|c|c|c|c}
\toprule
\rowcolor{gray!15}
\textbf{Backbone} & \textbf{Configuration} & \textbf{mAP} & \textbf{R-1} & \textbf{R-5} & \textbf{R-10} \\
\midrule
\multirow{5}{*}{TransReID}
  & Pretrained                & 37.1 & 39.7 & \textbf{79.0} & \textbf{89.2} \\
  & Finetuned                 & 37.4 & \underline{40.6} & 78.9 & 89.0 \\
  & Finetuned + Memory        & 37.5 & 39.5 & 77.9 & 88.5 \\
  & Finetuned + LoRA          & \underline{37.7} & \textbf{41.1} & \underline{78.9} & \underline{89.0} \\
  & Finetuned + Memory + LoRA & \textbf{38.1} & 39.9 & 78.1 & 88.4 \\
\midrule
\multirow{5}{*}{MagiV2}
  & Pretrained                & 50.8 & 57.1 & \underline{85.1} & \textbf{91.0} \\
  & Finetuned                 & 51.2 & \underline{57.2} & 85.1 & 90.8 \\
  & Finetuned + Memory        & \textbf{51.7} & 56.3 & 83.8 & 90.0 \\
  & Finetuned + LoRA          & 50.7 & \textbf{57.7} & \textbf{85.2} & \underline{91.0} \\
  & Finetuned + Memory + LoRA & \underline{51.7} & 56.3 & 83.9 & 90.1 \\
\midrule
\multirow{5}{*}{MagiV3}
  & Pretrained                & 40.7 & 47.7 & 82.5 & 90.6 \\
  & Finetuned                 & 41.3 & \textbf{48.7} & \textbf{83.2} & \textbf{90.7} \\
  & Finetuned + Memory        & \textbf{42.4} & 47.6 & 82.1 & 90.2 \\
  & Finetuned + LoRA          & 41.3 & \underline{48.7} & \underline{83.2} & \underline{90.7} \\
  & Finetuned + Memory + LoRA & \underline{42.3} & 47.7 & 82.4 & 90.2 \\
\midrule
\multirow{5}{*}{InstructReID}
  & Pretrained                & 36.7 & 40.2 & 78.8 & 88.6 \\
  & Finetuned                 & 37.8 & 42.5 & 79.3 & \underline{88.8} \\
  & Finetuned + Memory        & 38.0 & 41.3 & 78.0 & 88.1 \\
  & Finetuned + LoRA          & \underline{38.4} & \textbf{45.1} & \textbf{81.2} & \textbf{89.8} \\
  & Finetuned + Memory + LoRA & \textbf{40.1} & \underline{43.9} & \underline{79.9} & 88.5 \\
\midrule
\multirow{5}{*}{ReID5o}
  & Pretrained                & 37.8 & 43.6 & 80.2 & 89.2 \\
  & Finetuned                 & 38.7 & 44.8 & 80.3 & 89.3 \\
  & Finetuned + Memory        & 39.1 & 43.8 & 79.6 & 88.7 \\
  & Finetuned + LoRA          & \underline{39.1} & \underline{46.5} & \underline{81.3} & \textbf{89.9} \\
  & Finetuned + Memory + LoRA & \textbf{41.0} & \textbf{46.9} & \textbf{81.4} & \underline{89.8} \\
\bottomrule
\end{tabular}
\end{table}

\paragraph{Per-manga breakdown.}\label{app:per_manga} Aggregate \PopChars numbers can mask substantial per-series variation. Table~\ref{tab:suppl_per_manga} reports MagiV2's mAP and Rank-1 across the eight held-out series for the three configurations that bracket the headline. The per-series split shows the block's trade of Rank-1 for mAP in detail. \MeCha raises mAP on seven of the eight series, by 0.05 to 1.3 points, and lowers it only on Nisekoi, while Rank-1 falls on six of the eight, by up to 2.4 on Hunter $\times$ Hunter. Nisekoi is the only series that loses on both metrics, and Hunter $\times$ Hunter, which gives up the most Rank-1, gains 0.5 mAP while doing so, so the block redistributes similarity mass rather than adding it.

\begin{table}[!ht]
\centering \footnotesize \setlength{\tabcolsep}{5pt} \renewcommand{\arraystretch}{1.1} \caption{\textbf{Per-manga P1 results} for MagiV2 across three configurations, mean over three training runs and five seed draws. \#C is the character count. Best per series in \textbf{bold}.} \label{tab:suppl_per_manga}
\begin{tabular}{l|c|cc|cc|cc}
\toprule
\rowcolor{gray!15}
& & \multicolumn{2}{c|}{\textbf{Finetuned}} & \multicolumn{2}{c|}{\textbf{FT + Memory}} & \multicolumn{2}{c}{\textbf{FT + Mem + LoRA}} \\
\rowcolor{gray!15}
\multirow{-2}{*}{\textbf{Manga}} & \multirow{-2}{*}{\textbf{\#C}} & \textbf{mAP} & \textbf{R-1} & \textbf{mAP} & \textbf{R-1} & \textbf{mAP} & \textbf{R-1} \\
\midrule
Bakuman & 7 & 62.9 & \textbf{70.2} & \textbf{63.5} & 68.7 & \textbf{63.5} & \textbf{70.2} \\
Demon Slayer Kimetsu No Yaiba & 11 & 47.2 & \textbf{52.7} & \textbf{48.5} & 52.5 & 48.0 & 51.7 \\
Dr Stone & 4 & \textbf{63.8} & 65.6 & \textbf{63.8} & 65.0 & \textbf{63.8} & \textbf{66.0} \\
Hunter X Hunter & 12 & 40.7 & \textbf{48.8} & 41.2 & 46.5 & \textbf{41.6} & 45.4 \\
Kagurabachi & 6 & 41.2 & 51.0 & \textbf{42.0} & \textbf{52.4} & 41.7 & 52.3 \\
Nisekoi False Love & 10 & \textbf{50.8} & \textbf{52.2} & 50.6 & 50.1 & 50.6 & 51.2 \\
Oshi No Ko & 10 & 47.4 & 55.7 & \textbf{48.0} & \textbf{55.8} & \textbf{48.0} & 54.7 \\
Tokyo Ghoul & 10 & 55.6 & \textbf{61.2} & \textbf{56.3} & 59.4 & 56.0 & 59.2 \\
\bottomrule
\end{tabular}
\end{table}

\paragraph{Cross-corpus transfer.}\label{app:cross_dataset} The \PopChars headline numbers extend to two further evaluation corpora. Table~\ref{tab:manga109_results} scores the 27 held-out Manga109 volumes, starting from the released backbone and then adding \PopChars training, so the cost of transfer and the value of training can be read separately. The cross-protocol pattern from the main body persists. The manga-native backbones lead, and the gap between best and worst is wider than on \PopChars, so domain-specific pre-training matters more on the larger and more visually diverse corpus.

Training on \PopChars transfers. Against the same backbone with no \PopChars training at all, BNNeck fine-tuning is worth $+0.63$ to $+1.92$ P1 mAP on Manga109, positive on all five, and the best \MeCha configuration, which carries LoRA on four of the five, adds a further $+1.2$ to $+5.1$. The fine-tuning figure is larger than the $+0.28$ to $+1.10$ the same step is worth in domain (Table~\ref{tab:suppl_p1_full}), so none of the five pays for in-domain accuracy with transfer. Level is a different question from gain. Relative to its own \PopChars score each backbone loses 1.9 to 8.3 mAP here, except MagiV2, which gains 14.7. Cast size differs between the corpora, so those levels are not directly comparable and only the ordering by pre-training family is. This is cross-corpus transfer within Japanese comics, not out-of-domain evaluation.

On the Re:Verse benchmark over Re:Zero (Table~\ref{tab:reverse_results}) the same contrast is larger. The three vision-language models (Qwen2.5-VL-3B, InternVL3-14B, Ovis2-8B) score below 1.2\% character-identification accuracy. Every Re-ID encoder clears 30 mAP, and \MeCha with MagiV2 reaches 84.8. Re:Verse is therefore a useful sanity check that comic-character Re-ID is a real problem with its own structure, not a regime where image-language alignment alone is sufficient.

\begin{table}[t]
\centering
\footnotesize
\setlength{\tabcolsep}{4pt}
\renewcommand{\arraystretch}{1.02}
\caption{\textbf{Cross-corpus transfer to Manga109}, 27 held-out volumes (Appendix~\ref{app:reeval}). \emph{Pretrained} is the released backbone with no \PopChars training at all. The two rows under it are \PopChars-trained checkpoints scored zero-shot here, the no-memory baseline and then the best \MeCha configuration by P1 mAP ($\dagger$), so the first gap in each block is what BNNeck fine-tuning on another corpus is worth. P4 is identity Rank-1, with P2 given at Rank-1 beside it.}
\label{tab:manga109_results}
\resizebox{\linewidth}{!}{%
\begin{tabular}{l|l|cc|c|cc}
\toprule
\rowcolor{gray!15}
& & \multicolumn{2}{c|}{\textbf{P1: Closed-Set}} & \textbf{P2-R}$_{k=1}$ & \multicolumn{2}{c}{\textbf{Seq-R}$_{k=1}$, id.\ R-1} \\
\rowcolor{gray!15}
\multirow{-2}{*}{\textbf{Backbone}} & \multirow{-2}{*}{\textbf{Config}} & \textbf{mAP} & \textbf{R-1} & \textbf{mAP} & \textbf{P2} & \textbf{P4} \\
\midrule
TransReID & Pretrained & 28.4 & 35.9 & 26.6 & 12.2 & 10.5 \\
 & Finetuned & 29.1 & 36.5 & 26.9 & 12.5 & 10.2 \\
 & FT + Mem + LoRA$^\dagger$ & 30.3 & 37.1 & 27.9 & 13.3 & 10.7 \\
\midrule
MagiV2 & Pretrained & 65.3 & 76.6 & 63.8 & 50.4 & 45.7 \\
 & Finetuned & 65.9 & 77.4 & 63.6 & 50.5 & 45.7 \\
 & FT + Mem$^\dagger$ & 67.4 & 75.9 & 63.2 & 50.2 & 47.1 \\
\midrule
MagiV3 & Pretrained & 37.4 & 52.8 & 35.5 & 20.5 & 14.8 \\
 & Finetuned & 39.4 & 53.9 & 37.5 & 22.2 & 17.1 \\
 & FT + Mem + LoRA$^\dagger$ & 41.5 & 52.4 & 38.2 & 22.6 & 18.4 \\
\midrule
InstructReID & Pretrained & 28.6 & 39.5 & 26.6 & 12.7 & 9.7 \\
 & Finetuned & 30.4 & 41.1 & 28.1 & 13.9 & 10.6 \\
 & FT + Mem + LoRA$^\dagger$ & 35.5 & 46.3 & 32.3 & 17.6 & 14.3 \\
\midrule
ReID5o & Pretrained & 31.2 & 45.0 & 29.8 & 15.2 & 12.1 \\
 & Finetuned & 33.0 & 47.1 & 30.8 & 16.3 & 12.7 \\
 & FT + Mem + LoRA$^\dagger$ & 36.5 & 49.4 & 32.5 & 18.0 & 15.0 \\
\bottomrule
\end{tabular}%
}
\end{table}

\begin{table}[!ht]
\centering \footnotesize \setlength{\tabcolsep}{4pt} \renewcommand{\arraystretch}{1.02} \caption{\textbf{Re:Verse benchmark.} Character identification on Re:Zero. The VLM rows carry the character-identification accuracy published with the benchmark~\citep{baranwal2025reverse}. Every Re-ID row is our own harness (Appendix~\ref{app:reeval}), the mean over five seed draws, so the pretrained rows and the trained rows are directly comparable. \MeCha is worth $+0.1$ mAP here and costs 1.6 Rank-1, on a single series.} \label{tab:reverse_results}
\begin{tabular}{l|l|c|c|c}
\toprule
\rowcolor{gray!15}
& \textbf{Method} & \textbf{Char-ID Acc~(\%)} & \textbf{mAP} & \textbf{Rank-1} \\
\midrule
\multirow{3}{*}{\rotatebox[origin=c]{90}{\scriptsize VLMs}}
& Qwen2.5-VL-3B~\citep{baranwal2025reverse} & 1.11 & -- & -- \\
& InternVL3-14B~\citep{baranwal2025reverse} & 0.00 & -- & -- \\
& Ovis2-8B~\citep{baranwal2025reverse} & 0.85 & -- & -- \\
\midrule
\multirow{5}{*}{\rotatebox[origin=c]{90}{\scriptsize Re-ID}}
& TransReID (pre) & -- & 34.5 & 46.0 \\
& InstructReID (pre) & -- & 30.7 & 50.9 \\
& ReID5o (pre) & -- & 36.2 & 60.8 \\
& MagiV3 (pre) & -- & 48.2 & 73.0 \\
& MagiV2 (pre) & -- & \textbf{84.2} & \textbf{91.0} \\
\midrule
\rowcolor{gray!15}
\multirow{2}{*}{\rotatebox[origin=c]{90}{\scriptsize Ours}}
& MagiV2 (FT only) & -- & 84.7 & \textbf{91.7} \\
\rowcolor{gray!15}
& MagiV2 + \MeCha & -- & \textbf{84.8} & 90.1 \\
\bottomrule
\end{tabular}
\end{table}

\paragraph{Imperfect crops.}\label{app:crop_noise} All protocols operate on ground-truth detections, which isolates Re-ID from localisation. To measure what a detector would cost, we re-ran P1 and P2 under two families of synthetic perturbation on TransReID and MagiV2, each with and without \MeCha, over three seed draws and macro over the 8 held-out series. The first family displaces the bounding box before cropping, either shifting the centre by 10, 20 or 30 percent of the box size in a random direction or scaling the box to $0.7\times$ or $1.3\times$ (Table~\ref{tab:suppl_boxnoise}). The second degrades the pixels inside a correct box, by brightness and contrast jitter, Gaussian blur, or occluding a random rectangle (Table~\ref{tab:suppl_pixelnoise}). Cells after the first row are changes from the clean condition in the same column.

Box placement error is benign and pixel degradation is not. A 30 percent centre shift costs TransReID 0.7 P1 mAP and MagiV2 2.4, and a $1.3\times$ loose box costs 0.8 and 2.8, while mild displacement is neutral or slightly positive and acts as augmentation. Inside a correct box the two backbones separate. Brightness and contrast jitter is almost free on both. A 30 percent occlusion costs MagiV2 6.9 P1 mAP against TransReID's 1.3, and a $\sigma{=}4$ blur costs MagiV2 4.9 while leaving TransReID unmoved at $+0.2$. MagiV2's comic-native features carry more of the signal that blur destroys, so it has more to lose. A clean-crop ranking is therefore not a deployment ranking.

\MeCha earns slightly more as the crop degrades, on MagiV2 only. On MagiV2 it is worth $+0.55$ P1 mAP on clean crops and $+0.66$ averaged over the eleven perturbed conditions, peaking at $+0.91$ under a 20 percent shift. On TransReID, where it is worth $+0.12$ clean, the same average is $+0.11$. A context-carrying representation helps most where the crop itself carries least, but the effect is a tenth of a point.

\begin{table}[!ht]
\centering \footnotesize \setlength{\tabcolsep}{5pt} \renewcommand{\arraystretch}{1.1} \caption{\textbf{Bounding-box displacement.} P1 and P2 Seq-R at $k{=}1$, three seed draws, macro over the 8 test series. The first row is absolute mAP and every later row is the change from it in the same column. IoU is the mean overlap of the displaced box with the true one.} \label{tab:suppl_boxnoise}
\begin{tabular}{l|c|cc|cc|cc|cc}
\toprule
\rowcolor{gray!15}
& & \multicolumn{2}{c|}{\textbf{TransReID FT}} & \multicolumn{2}{c|}{\textbf{TransReID +Mem}} & \multicolumn{2}{c|}{\textbf{MagiV2 FT}} & \multicolumn{2}{c}{\textbf{MagiV2 +Mem}} \\
\rowcolor{gray!15}
\multirow{-2}{*}{\textbf{Box condition}} & \multirow{-2}{*}{\textbf{IoU}} & \textbf{P1} & \textbf{P2} & \textbf{P1} & \textbf{P2} & \textbf{P1} & \textbf{P2} & \textbf{P1} & \textbf{P2} \\
\midrule
clean box & 1.00 & 37.4 & 38.6 & 37.5 & 38.4 & 51.2 & 54.4 & 51.7 & 54.5 \\
\midrule
shift 10\% & 0.83 & +0.1 & +0.2 & +0.2 & -0.1 & -0.0 & +0.4 & +0.2 & +0.0 \\
shift 20\% & 0.69 & -0.2 & -0.3 & -0.3 & -0.4 & -0.9 & -1.4 & -0.5 & -1.4 \\
shift 30\% & 0.58 & -0.7 & -0.4 & -0.8 & -0.2 & -2.4 & -2.7 & -2.1 & -2.7 \\
tight $0.7\times$ & 0.49 & +0.6 & +1.5 & +0.6 & +1.6 & -0.6 & -0.2 & -0.5 & -0.7 \\
loose $1.3\times$ & 0.59 & -0.8 & -1.8 & -0.8 & -1.7 & -2.8 & -1.6 & -2.6 & -1.7 \\
\bottomrule
\end{tabular}
\end{table}

\begin{table}[!ht]
\centering \footnotesize \setlength{\tabcolsep}{5pt} \renewcommand{\arraystretch}{1.1} \caption{\textbf{Pixel corruption} (same setting as Table~\ref{tab:suppl_boxnoise}). Blur and occlusion, not box placement, are what separate the two backbones.} \label{tab:suppl_pixelnoise}
\begin{tabular}{l|cc|cc|cc|cc}
\toprule
\rowcolor{gray!15}
& \multicolumn{2}{c|}{\textbf{TransReID FT}} & \multicolumn{2}{c|}{\textbf{TransReID +Mem}} & \multicolumn{2}{c|}{\textbf{MagiV2 FT}} & \multicolumn{2}{c}{\textbf{MagiV2 +Mem}} \\
\rowcolor{gray!15}
\multirow{-2}{*}{\textbf{Pixel condition}} & \textbf{P1} & \textbf{P2} & \textbf{P1} & \textbf{P2} & \textbf{P1} & \textbf{P2} & \textbf{P1} & \textbf{P2} \\
\midrule
clean pixels & 37.4 & 38.6 & 37.5 & 38.4 & 51.2 & 54.4 & 51.7 & 54.5 \\
\midrule
jitter 10\% & +0.0 & +0.3 & +0.1 & -0.3 & -0.1 & +0.1 & -0.0 & +0.0 \\
jitter 20\% & -0.2 & +0.1 & -0.2 & -0.5 & -0.5 & -0.3 & -0.5 & -0.4 \\
blur $\sigma{=}2$ & +0.2 & +0.6 & +0.1 & +0.3 & -1.2 & -1.2 & -1.2 & -1.3 \\
blur $\sigma{=}4$ & +0.2 & +0.4 & -0.1 & -0.7 & -4.9 & -5.4 & -5.0 & -6.3 \\
occlusion 15\% & -0.5 & +0.3 & -0.3 & +0.5 & -3.1 & -2.0 & -3.0 & -1.9 \\
occlusion 30\% & -1.3 & -1.2 & -1.1 & -1.4 & -6.9 & -7.6 & -7.1 & -7.0 \\
\bottomrule
\end{tabular}
\end{table}

\FloatBarrier
\subsection{Identity emergence (P3)} \label{app:p3}

P3 evaluates identity \emph{emergence}. The system encounters an empty gallery and must instantiate clusters as new characters appear, deciding for each new observation whether it joins an existing cluster (similarity above $\tau_{\mathrm{nov}}$) or seeds a new one. We evaluate on full per-series streams, every crop of each held-out series in reading order. We report predicted cluster counts, Purity, Hungarian-matched accuracy, and ARI alongside NMI, because Purity alone is gameable by fragmentation.

\textbf{Alternative decision rules.} P3 defines the task and admits any decision rule. The fixed threshold is the reference instantiation that compares every representation under one criterion. The rules below were evaluated to test whether per-identity calibration changes the picture. They are not proposed methods or baselines of the framework, no conclusion in the paper depends on their ranking, and the protocol and its reported results do not change with the rule.

Table~\ref{tab:suppl_p3_rules} evaluates four alternatives on the same full-stream features and reading-order streams, namely variance-adaptive per-cluster thresholds ($\tau_c=\mu_c-\lambda\sigma_c$ over join-time similarities), density-aware thresholds (median core similarity with a minimum core size $n_{\min}$), cohesion-relative thresholds (proportional to the running mean similarity), and graph community detection (Louvain over a mutual-$k$NN graph with reading-order edges). Every alternative relocates the operating point along the same purity-versus-fragmentation frontier instead of moving above it. Variance-adaptive, density-aware and cohesion-relative each raise Purity, by 2.1 to 9.3 points, and pay for it in fragmentation, instantiating 1.2 to 3.8 times as many clusters as the fixed rule. All six of those cells lose both Hungarian accuracy and ARI. Graph community detection moves the other way and merges instead, to 0.23 of the fixed rule's clusters on TransReID and 0.63 on MagiV2. On MagiV2 that costs 13.6 points of ARI. On TransReID it is the one rule anywhere in the table that edges past the fixed rule, 2.6 ARI against 2.0, and it pays 9.2 points of Purity for it. Per-identity calibration therefore reveals nothing a global threshold hides, and what binds is the representation's geometry.

P3 begins from an empty gallery, so when an identity first emerges there is no per-identity variance or density to calibrate against. A shared criterion is the only rule available at the moment the decision has to be made.

Two details of Table~\ref{tab:suppl_p3_rules} matter for reading it. All cluster counts are macro averages per series, against 8.8 identities per series, rather than corpus totals against the 70 held-out characters. On that basis the graph rule's 21.8 and 24.6 clusters are roughly 2.5-fold over-segmentation rather than under-segmentation. The graph is also built on \emph{mutual} $k$-nearest neighbours, so an edge requires reciprocity and a rare crop cannot bridge identities on its own. Nor does the long tail explain the density rule's fragmentation. Characters with four or fewer crops are 11 of the 70 held-out characters and hold 24 crops in total, 3.0 crops per series or 0.59 percent of the data, so even if every one became a singleton it would add about 3 clusters per series against an observed excess of 85 on TransReID and 110 on MagiV2 over the fixed rule.

Density-aware and graph-based rules are nonetheless the most promising next algorithms for identity emergence. Sweeping each alternative over its own six-point grid, against the fixed rule over twelve thresholds, leaves the picture as it is. On MagiV2 all 24 alternative settings sit below the fixed rule at a matched cluster count, by 1.7 ARI or more, and the best ARI of the whole sweep is the fixed rule's own, 26.4 at $\tau_{\mathrm{nov}}=0.45$. On TransReID no setting clears the fixed rule by more than 0.03 ARI where the two span the same cluster counts, and beyond that range graph community detection, at 14.0 and 10.2 clusters, reaches 2.6 and 3.4 against the fixed rule's best of 3.0, all near chance.

\begin{table}[!ht]
\centering \footnotesize \setlength{\tabcolsep}{5pt} \renewcommand{\arraystretch}{1.1} \caption{\textbf{P3 decision rules on full per-series streams} (finetuned TransReID and MagiV2 from the first training run, \PopChars test series, macro over 8 series). Best ARI per backbone in \textbf{bold}. No alternative improves on the fixed rule at a comparable cluster count. On MagiV2 every rule is below it, and on TransReID graph community detection edges past it only by collapsing to a quarter of the clusters.} \label{tab:suppl_p3_rules}
\begin{tabular}{l|l|ccccc}
\toprule
\rowcolor{gray!15}
\textbf{Backbone} & \textbf{Rule} & \textbf{\#clusters} & \textbf{Purity} & \textbf{Hung.\ Acc} & \textbf{ARI} & \textbf{NMI} \\
\midrule
TransReID & fixed $\tau_{\mathrm{nov}}=0.55$ & 95.8 & 60.2 & 15.1 & 2.0 & 21.8 \\
 & variance-adaptive & 111.2 & 62.3 & 13.8 & 1.6 & 22.9 \\
 & density-aware & 180.6 & 69.5 & 8.6 & 0.9 & 27.8 \\
 & cohesion-relative & 144.0 & 65.9 & 11.4 & 1.4 & 25.8 \\
 & graph community detection & 21.8 & 51.0 & 17.5 & \textbf{2.6} & 13.0 \\
\midrule
MagiV2 & fixed $\tau_{\mathrm{nov}}=0.55$ & 38.9 & 69.6 & 46.6 & \textbf{24.0} & 34.1 \\
 & variance-adaptive & 62.1 & 72.4 & 39.3 & 18.4 & 34.7 \\
 & density-aware & 149.2 & 77.8 & 15.5 & 3.8 & 34.0 \\
 & cohesion-relative & 89.9 & 75.6 & 30.9 & 13.6 & 35.7 \\
 & graph community detection & 24.6 & 65.4 & 29.2 & 10.4 & 29.2 \\
\bottomrule
\end{tabular}
\end{table}

\textbf{Identity emergence under the fixed rule.} Table~\ref{tab:p3_main} gives the full picture behind Section~\ref{subsec:emergence}. Under the fixed rule finetuned TransReID and MagiV2 reach Hungarian-matched accuracy of 15.1 and 46.6 and ARI of 2.0 and 24.0, and a looser threshold raises their Purity to 93.0 and 84.3 only by fragmenting each identity into dozens of clusters, which drops ARI to 0.2 and 9.8.


\textbf{Maintenance-tuned representations on P3.} Table~\ref{tab:suppl_p3} compares each finetuned baseline with the same backbone trained alongside \MeCha on full per-series streams, across all five backbones. Training alongside the block costs 0.9 to 2.2 points of Purity on four of five backbones and returns 0.1 to 0.8 of ARI on four of five. Both shifts are small against the differences between backbones, whose ARI ranges from 1.8 to 24.0, so maintenance training leaves emergence where the backbone put it. \Recog measures maintenance and emergence separately so that a trade-off between them, where one appears, is visible rather than averaged away.

\begin{table}[!ht]
\centering \footnotesize \setlength{\tabcolsep}{5pt} \renewcommand{\arraystretch}{1.2} \caption{\textbf{P3 on full per-series streams, all five backbones}, fixed rule at $\tau_{\mathrm{nov}}=0.55$, macro over the 8 held-out \PopChars series against 8.8 identities per series, first training run. Every crop of every series is streamed in reading order. P3 has no gallery to initialise memory from, so the block is bypassed and the right-hand columns score the BNNeck trained alongside it.} \label{tab:suppl_p3}
\begin{tabular}{l|ccccc|ccccc}
\toprule
\rowcolor{gray!15}
& \multicolumn{5}{c|}{\textbf{Finetuned}} & \multicolumn{5}{c}{\textbf{Trained with \MeCha}} \\
\rowcolor{gray!15}
\textbf{Backbone} & \textbf{\#cl.} & \textbf{Pur.} & \textbf{Hung.} & \textbf{ARI} & \textbf{NMI} & \textbf{\#cl.} & \textbf{Pur.} & \textbf{Hung.} & \textbf{ARI} & \textbf{NMI} \\
\midrule
TransReID & 95.8 & 60.2 & 15.1 & 2.0 & 21.8 & 90.1 & 59.3 & 16.3 & 2.2 & 21.3 \\
MagiV2 & 38.9 & 69.6 & 46.6 & 24.0 & 34.1 & 39.9 & 69.9 & 47.4 & 24.5 & 34.5 \\
MagiV3 & 79.9 & 65.2 & 20.7 & 5.9 & 26.0 & 69.0 & 64.1 & 23.0 & 6.7 & 24.6 \\
InstructReID & 278.8 & 81.0 & 10.5 & 2.1 & 33.3 & 260.6 & 78.8 & 11.0 & 2.2 & 32.2 \\
ReID5o & 213.8 & 76.0 & 10.6 & 1.8 & 31.2 & 199.1 & 74.3 & 10.5 & 1.8 & 30.4 \\
\bottomrule
\end{tabular}
\end{table}

\begin{figure}[!ht]
\centering \includegraphics[width=\linewidth]{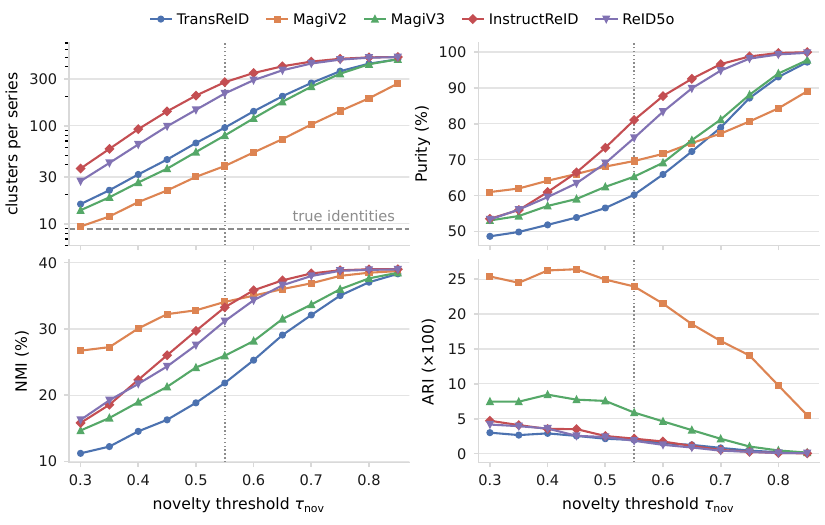} \caption{\textbf{P3 novelty-threshold sensitivity} on full per-series streams, showing predicted clusters per series (log scale, with the 8.8 true identities dashed), Purity, NMI and ARI against $\tau_{\mathrm{nov}}$ for the five finetuned backbones, first training run, macro over the 8 test series. The dotted line is the reference $\tau_{\mathrm{nov}}=0.55$. Purity and NMI rise as the stream fragments, ARI falls or peaks early, and MagiV2 leads on ARI at every threshold.} \label{fig:suppl_threshold}
\end{figure}

\paragraph{Novelty threshold $\tau_{\mathrm{nov}}$.}\label{app:hyperparam} P3 admits one free parameter, the cosine similarity above which a crop joins a known cluster rather than opening a new one. Figure~\ref{fig:suppl_threshold} sweeps it from 0.30 to 0.85 for the five finetuned backbones on full per-series streams and reports all four quantities P3 is scored by. Purity and NMI rise with the threshold on every backbone, with no plateau, and they rise because the stream fragments. At 0.85 every backbone but MagiV2 opens more than 470 clusters for 8.8 identities per series and reaches a Purity of 97 or more. ARI and the cluster count expose it. ARI falls as the threshold rises on the three non-manga backbones and peaks at 0.40 to 0.45 on the two manga-native ones, and only MagiV2, at the lowest threshold, opens about as many clusters as there are identities. The protocol keeps its reference value, $\tau_{\mathrm{nov}}=0.55$, for every representation, so that the comparison is between embeddings rather than between tunings, and the sweep shows the comparison does not hinge on it, since at every threshold MagiV2's ARI is at least three times that of any other backbone. Purity and NMI alone do not identify an operating point, which is why P3 reports ARI and the predicted cluster count beside them (Table~\ref{tab:p3_main}).

\FloatBarrier
\subsection{Gallery growth (P4)} \label{app:p4_dynamics}

\begin{table}[t]
\centering
\footnotesize
\setlength{\tabcolsep}{4pt}
\renewcommand{\arraystretch}{1.02}
\caption{\textbf{P4 under its three update policies} at $k=1$ and $B_{\max}=50$ on the 8 held-out \PopChars test series, identity Rank-1, mean over three training runs and five seed draws. \emph{Static} is the unchanged P2 gallery, \emph{predicted} adds each query under its top-1 match (the protocol's rule) and \emph{oracle} adds each query under its true character. \emph{Wrong-app.} is the fraction of added entries that carry the wrong character and \emph{contam.} the mislabelled fraction of the grown gallery at stream end, both under the predicted policy. \emph{Finetuned} trains a BNNeck head on the frozen backbone with no LoRA, and \emph{FT + Mem} adds \MeCha to it. The three policies share one seed draw per run.}
\label{tab:p4_oracle}
\resizebox{\linewidth}{!}{%
\begin{tabular}{l|l|ccc|cc|ccc|cc}
\toprule
\rowcolor{gray!15}
& & \multicolumn{5}{c|}{\textbf{Seq-R} (random seeding)} & \multicolumn{5}{c}{\textbf{Seq-T} (chronological seeding)} \\
\rowcolor{gray!15}
\multirow{-2}{*}{\textbf{Backbone}} & \multirow{-2}{*}{\textbf{Config}} & \textbf{Static} & \textbf{Pred.} & \textbf{Oracle} & \textbf{Wrong-app.} & \textbf{Contam.} & \textbf{Static} & \textbf{Pred.} & \textbf{Oracle} & \textbf{Wrong-app.} & \textbf{Contam.} \\
\midrule
TransReID & Finetuned & 17.4 & 15.0 & 41.7 & 85.0 & 84.2 & 12.8 & 12.6 & 42.2 & 87.4 & 83.3 \\
 & FT + Mem & 17.2 & 15.7 & 41.4 & 84.3 & 82.8 & 13.2 & 13.6 & 41.8 & 86.4 & 84.0 \\
\midrule
MagiV2 & Finetuned & 36.8 & 36.3 & 59.1 & 63.7 & 67.8 & 36.6 & 37.9 & 59.0 & 62.1 & 69.8 \\
 & FT + Mem & 37.2 & 35.9 & 59.8 & 64.1 & 68.7 & 35.3 & 36.9 & 60.4 & 63.1 & 68.3 \\
\midrule
MagiV3 & Finetuned & 22.7 & 19.9 & 49.9 & 80.1 & 80.0 & 17.4 & 17.7 & 50.2 & 82.3 & 82.0 \\
 & FT + Mem & 23.5 & 19.6 & 50.5 & 80.4 & 80.9 & 17.7 & 17.5 & 50.7 & 82.5 & 80.9 \\
\midrule
InstructReID & Finetuned & 17.5 & 14.5 & 42.6 & 85.5 & 84.1 & 13.3 & 10.6 & 42.9 & 89.4 & 87.4 \\
 & FT + Mem & 17.8 & 16.1 & 42.7 & 83.9 & 83.7 & 14.1 & 11.3 & 43.0 & 88.7 & 85.4 \\
\midrule
ReID5o & Finetuned & 20.8 & 17.9 & 45.6 & 82.1 & 82.1 & 12.8 & 14.4 & 45.6 & 85.6 & 84.3 \\
 & FT + Mem & 21.1 & 18.5 & 46.3 & 81.5 & 81.7 & 12.6 & 14.7 & 46.8 & 85.3 & 85.3 \\
\bottomrule
\end{tabular}%
}
\end{table}

Table~\ref{tab:p4_oracle} decomposes P4 at $k=1$ and the protocol's operating point $B_{\max}=50$ into a static gallery, growth by the model's own top-1 match, and growth under every query's true character. Under random seeding the oracle sits 22.3 to 27.1 identity Rank-1 above the static gallery without \MeCha and 22.7 to 27.0 with it, while predicted growth is below the static gallery in all ten cells, by 0.5 to 4.0. Under chronological seeding the oracle's margin is larger, 22.5 to 34.2, and predicted growth splits, with six of the ten cells positive and none by more than 2.2. Contamination is the reason. Between 62 and 89 percent of appended entries carry the wrong identity, and at stream end between 68 and 87 percent of the grown gallery is mislabelled.

\begin{figure}[!ht]
\centering \includegraphics[width=\linewidth]{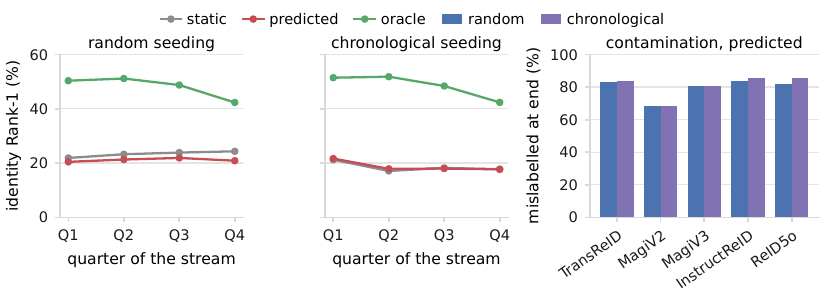} \caption{\textbf{P4 over the stream} at $k=1$ and $B_{\max}=50$ with \MeCha, mean over three training runs, five seed draws and the 8 test series. The left and centre panels give identity Rank-1 in each quarter of the stream, pooled over the five backbones, for the static gallery, growth by the model's own top-1 (predicted) and growth by the true label (oracle). The right panel gives the share of the grown gallery that is mislabelled at the end of the stream under predicted growth.} \label{fig:p4_drift}
\end{figure}

Two further statistics complete the picture (Figure~\ref{fig:p4_drift}). First, the aggregate is not hiding a runaway series. With \MeCha under random seeding the worst single-series decline from the static gallery is 13.5 identity Rank-1 (MagiV3 on Dr Stone) against macro declines of 1.3 to 4.0, and under chronological seeding the per-series spread is wider in both directions, from 9.4 down to 18.6 up. Second, stream position is the sharper diagnostic of drift. Pooled over the five backbones with \MeCha under random seeding, the static gallery is flat across the four quartiles of the stream (21.9, 23.3, 23.9 and 24.3 identity Rank-1), predicted growth stays between 1.4 and 3.5 points below it in every quartile, and the oracle's lead narrows from 28.4 points in the first quartile to 18.0 in the last, between 16.4 and 20.3 across backbones. Correct updates therefore buy the most where the gallery has seen the least, and contaminated ones do not snowball, since the deficit they open widens by about two points over an entire stream. P4 is a protocol and not a method with an unguarded update rule. Self-updating galleries are what deployed systems do, and P4 makes their benefit and their contamination measurable together instead of assuming them away.

\textbf{Buffer size.} Table~\ref{tab:suppl_bmax} sweeps $B_{\max}$ over $\{0, 5, 10, 25, 50, 100, \infty\}$ at $k=1$. The cap bounds only the replaceable grown buffer and leaves protected seeds untouched. Across the grown galleries, $B_{\max}\geq 5$, MagiV2 moves by 4.1 identity Rank-1 under chronological seeding and 1.2 under random, TransReID by 1.8 and 1.1, and beyond $B_{\max}=25$ no configuration moves by more than 2.0. Under random seeding every cap leaves growth below the static gallery on both backbones, and under chronological seeding the two cross as the cap loosens without growth reaching the oracle's range, so no qualitative conclusion depends on the cap. With this sweep every free parameter of the protocol suite has a sensitivity study, the seed count $k$ over 1 to 5 throughout, the novelty threshold and the alternative rules in Appendix~\ref{app:p3}, and $B_{\max}$ here.

\begin{table}[!ht]
\centering \footnotesize \setlength{\tabcolsep}{5pt} \renewcommand{\arraystretch}{1.1} \caption{\textbf{Sensitivity of P4 to the buffer cap $B_{\max}$} (identity Rank-1 at $k=1$ on the 8 held-out \PopChars series, mean over three training runs). $B_{\max}=0$ is the static P2 gallery. No cap turns growth by predicted labels into a gain over that gallery. Under random seeding every cap is below it, and under chronological seeding MagiV2 recovers about two points as the cap loosens without reaching the oracle's range. The cap is therefore not the free parameter that decides whether growth pays. Acceptance is (Section~\ref{sec:vouch}).} \label{tab:suppl_bmax}
\begin{tabular}{l|c|cccccc}
\toprule
\rowcolor{gray!15}
\textbf{Configuration} & \textbf{0 (=P2)} & \textbf{5} & \textbf{10} & \textbf{25} & \textbf{50} & \textbf{100} & \textbf{unbounded} \\
\midrule
TransReID, chronological  & 13.2 & 13.2 & 14.2 & 13.0 & 13.6 & 14.6 & 14.9 \\
TransReID, random         & 17.2 & 16.1 & 16.4 & 15.9 & 15.7 & 15.4 & 15.4 \\
MagiV2, chronological     & 35.3 & 35.8 & 33.4 & 35.5 & 36.9 & 37.5 & 37.4 \\
MagiV2, random            & 37.2 & 35.4 & 34.9 & 35.9 & 35.9 & 36.1 & 36.1 \\
\bottomrule
\end{tabular}
\end{table}

\textbf{The acceptance gap does not shrink when the crop degrades.} Re-running P4 under the eleven conditions of Appendix~\ref{app:crop_noise} puts the oracle between 19.9 and 24.1 identity Rank-1 above the static gallery on TransReID and between 21.0 and 23.6 on MagiV2, against 24.3 and 22.2 on clean crops. Growth by the model's own top-1 stays below the static gallery in all eleven conditions on both finetuned backbones. The gap is therefore not an artefact of evaluating on ground-truth crops, because degrading the input moves the static gallery and the oracle together, and leaves the distance between them almost exactly where it was. Whatever a detector would cost this system, it would not cost it the headroom that Section~\ref{sec:vouch} is about.

\paragraph{Four attempts at chronological seeding.}\label{app:seqt} Section~\ref{subsec:seqt} says what makes chronological seeding hard, and four mechanisms were measured against it before the binding of Section~\ref{sec:binding}. Each falls short for its own reason, and together they point to what binding supplies, a correct reference near the query.

\textbf{A running mean grown by the model's own top-1.} An exponential prototype with anchored seeds is $+10.13$ identity Rank-1 on MagiV2 under chronological seeding ($p=0.009$, 8 of 8 series). It ranks 14th of 62 arms on the development series, so the configuration was selected on the test set. Under selection on development the chosen arm is positive on one of five backbones, not significantly, and significantly negative on two.

\textbf{Restricting the candidate identities by recency.} Ranking each query only against identities seen in the last twenty crops is worth $+3.4$ to $+5.7$ identity Rank-1 on four of five backbones under random seeding on \PopChars, and $+5.5$ to $+8.0$ on the same four across 27 held-out Manga109 volumes, where it costs MagiV2 $7.21$ against a static gallery already at 64.8. It obeys a decomposition of the same shape as Equation~\ref{eq:commit}. With $\ell$ the share of queries whose character survives the restriction, $r^{+}$ and $a^{+}$ the restricted and unrestricted accuracy on those queries, and $a^{-}$ the unrestricted accuracy on the rest, all of which the restriction gets wrong, $\Delta=\ell\,(r^{+}-a^{+})-(1-\ell)\,a^{-}$ exactly. The discarded queries are typical, $a^{-}=1.04\,a-5.0$ ($r=0.99$), so the cost term grows with the gallery's accuracy while the benefit term shows no trend, which is why the restriction that helps four galleries costs the strongest one. Under chronological seeding on \PopChars all five are positive and none significantly. The restriction is built from the system's own predictions, which are worse under chronological seeding, so the mechanism is weakest exactly where its headroom is largest.

\textbf{Pooling the causal window before matching.} Averaging a query with the crops within $0.8$ cosine of it in a 40-crop window is worth $+0.97$ to $+4.66$ on four of five backbones over 27 Manga109 volumes, with no anchor, growth or label, and group purity governs the sign ($r=+0.54$ over 20 cells). Over ten matched pairs the gain is $0.27$ \emph{lower} under chronological seeding than random, so the effect is a property of the corpus and not of the regime. Query expansion never needed an anchor, so anchor scarcity cannot hurt it, and it cannot address what Section~\ref{subsec:seqt} identifies.

\textbf{Pricing an append against the remaining stream.} An append at $t$ answers only queries after $t$, so its bar is the static gallery's accuracy over the remainder rather than over the whole stream. Measured per quarter under chronological seeding, the opening quarter's margin $p_{\mathrm{eff}}-a^{+}$ is $+0.020$ and the same margin priced forward is $+0.058$, against $-0.005$ and $-0.010$ under random seeding, so the correction is eight times larger where the static gallery is not flat. Acting on it by appending only while the stream position is below a threshold beats unrestricted growth on five of five backbones under random seeding and three of five under chronological, so the gain is from appending less rather than from pricing better. The correction is real and bounded at $+3.6$ identity Rank-1, because it lowers a bar without naming the crop that clears it.

The four share one shape, and the distance curve of Section~\ref{subsec:seqt} states it. Chronological seeding puts a query $0.407$ of the stream from its nearest own-identity seed against $0.094$ under random seeding, and identity Rank-1 falls at $-0.149$ per unit of that distance, which predicts 4.7 of the 5.1-point gap between the regimes. Re-weighting the chronological queries onto the random distance distribution takes the regime gap from $-5.1$ to $+0.2$ pooled and to between $-1.6$ and $+2.1$ per backbone, so the deficit is that one scalar. A restriction, a pooling and a forward price each change which identities compete or what an append costs, and a running mean changes what an entry contains. None moves a correct reference nearer the query. The only signal measured here that arrives independently of appearance is a name in a dialogue bubble, which clears the break-even on two of eight series while averaging below it, as measured below. Gating such an anchor on its own measured precision is a natural next step.

\textbf{The binding's own terms.} For each backbone binding with its own embedding at its selected threshold, the terms of Equation~\ref{eq:commit} are measured directly under chronological seeding at $k=5$. The binding captures 70 to 77 percent of the queries. Its $p_{\mathrm{eff}}$ is 24.0 to 28.2 percent on the four weaker backbones and 49.8 on MagiV2, within 4.4 points of the binding's own precision, and $a^{+}$ lies within 1.3 points of the static gallery's average accuracy, 20.0 to 38.7 against 20.4 to 37.4. The product $c\,(p_{\mathrm{eff}}-a^{+})$ reproduces the measured gains of 1.73 to 8.46 exactly. With most of the stream captured, the captured queries are close to the whole stream, which is why Section~\ref{subsec:ceiling} can compare a binding's precision with the gallery's average. For the shared binder on Manga109, whose additions are right 47.8 percent of the time, the static galleries under chronological seeding start at 15.5, 18.7, 22.9 and 30.3 on TransReID, InstructReID, ReID5o and MagiV3, which gain 10.32, 10.34, 10.35 and 8.00, and at 57.5 on MagiV2, which loses 7.44.

\paragraph{Four pairwise acceptance rules.}\label{app:acceptance} Four pairwise rules were measured on the same streams before the constraint was adopted. A confidence threshold on the top-1 cosine separates nothing, because on every backbone the top-1 cosine already exceeds 0.7 for essentially every query. A margin threshold, the top-1 cosine minus the best cosine to any other identity, is worth $+3.62$ identity Rank-1 on MagiV2 and is negative on MagiV3 and TransReID. A causal mutual-neighbour test, admitting a crop only when it is among the winning exemplar's $k$ nearest neighbours over the crops already delivered, is at or below the static gallery at every $k$ on both comic-native backbones. Averaging a query with the delivered members of its own page group before matching, which is neighbour aggregation restricted to the page, is $-6.42$ and $-4.40$.

\paragraph{Dialogue does not anchor identities across pages.} Every anchor in this paper is page-local, which limits how much of the stream commitment can reach. The obvious escape is dialogue. MagiV2 reads text boxes and associates each to its likely speaker, \PopChars\ names its identities, and a name spoken in a bubble is a label that arrives independently of appearance, on whatever page it falls. We measured it before building on it. Across the 8 test series, 385 bubbles contain a character name, and three readings of what the name refers to were scored against ground truth, namely the associated speaker (16.6\,\% correct over 331 firings), another character on the page as the addressee in the vocative case (29.1\,\% over 330) and the character nearest the speaker (15.8\,\% over 330). The vocative reading is the best of the three by 12.5 points, which confirms the linguistics, since a name in a bubble is usually spoken \emph{to} its bearer. No rule reaches the 40.9\,\% break-even \emph{on average}, but the addressee rule is bimodal across series and clears it on two of the eight, Bakuman at $30/36$ and Kagurabachi at $21/37$, both significant against the break-even and against 15.3\,\% on Nisekoi, which alone supplies 40\,\% of all firings. The aggregate therefore understates what the anchor is worth where it works. Page-local grouping, and the binding of Section~\ref{sec:binding} that links page groups across pages, remain the signals that clear the bar.

\paragraph{Where \Vouch turns on.} Commitment fires at exactly the rate at which a seed lands in the query's own page group, 2.5\% at $k=1$ and 9.5\% at $k=5$ under random seeding and 1.5\% and 2.3\% under chronological seeding. This is a property of the constraint's page locality rather than of the gallery's contents, because a crop committed from page $p-1$ belongs to that page's groups and never to page $p$'s. It is also why a gallery that tracks recent appearances does not help the rule, because recency changes which crops are in the gallery, not which page they were drawn on.

\FloatBarrier

\section{Broader impacts, ethics and reproducibility} \label{app:impact}

\textbf{Broader impacts and ethics.} The corpora consist of fictional manga characters, and all crops derive from published works used under their research licences, \PopChars from the publicly released PopManga annotations and Manga109 under its academic licence, of which we redistribute only a conversion script. No human subjects, personal data or real-person identities are involved. The methods evaluated here should not be deployed for real-person identification or surveillance, because the regularities that make comic-character Re-ID tractable, such as consistent character designs and controlled rendering, do not transfer to human-identity settings, and success here is no evidence of capability there.

\textbf{Reproducibility.} The protocols are specified in Section~\ref{subsec:protocols} and \MeCha in Appendices~\ref{app:mecha_arch} to~\ref{app:two_pass}, and Appendix~\ref{app:datasets} states how every reported metric is reproduced from the released harness. Code and the project page is \url{https://re-cognize.vercel.app}.


\end{document}